\RequirePackage{letltxmacro}
\documentclass[pdflatex,sn-mathphys]{sn-jnl}

\jyear{2021}%
\usepackage[utf8]{inputenc}
\usepackage[T5]{fontenc}
\usepackage{tabulary}
\usepackage{latexsym}
\usepackage{float}
\usepackage{makecell}
\usepackage{array,booktabs,arydshln,xcolor}
\usepackage{subcaption,booktabs}
\usepackage{microtype}
\usepackage{multirow}
\usepackage{multicol}
\usepackage{threeparttable} 
\usepackage{amsmath}
\usepackage{tablefootnote}
\usepackage{pifont}
\usepackage{float}
\usepackage{caption}
\usepackage{placeins}
\usepackage{array}
\usepackage{textcomp}
\usepackage{lmodern}
\usepackage{graphicx}
\usepackage{soul,color}
\usepackage{xurl}
\usepackage{amssymb}
\usepackage{pifont}
\newcommand{\cmark}{\ding{51}}%
\newcommand{\xmark}{\ding{55}}%
\theoremstyle{thmstyleone}%
\theoremstyle{thmstyletwo}%

\theoremstyle{thmstylethree}%

\begin{document}

\title[Hate and Offensive Detection using PhoBERT-CNN and Streaming Data]{Vietnamese Hate and Offensive Detection using PhoBERT-CNN and Social Media Streaming Data}



\author*[1,2]{\fnm{Khanh} \sur{Q. Tran}}\email{18520908@gm.uit.edu.vn}

\author[1,2]{\fnm{An} \sur{T. Nguyen}}\email{18520434@gm.uit.edu.vn}

\author[1,2]{\fnm{Phu} \sur{Gia Hoang}}\email{19520215@gm.uit.edu.vn}

\author[1,2]{\fnm{Canh} \sur{Duc Luu}}\email{19521272@gm.uit.edu.vn}

\author[1,2]{\fnm{Trong-Hop} \sur{Do}}\email{hopdt@uit.edu.vn}

\author[1,2]{\fnm{Kiet} \sur{Van Nguyen}}\email{kietnv@uit.edu.vn}

\affil*[1]{\orgname{University of Information Technology}, \orgaddress{ \city{Ho Chi Minh City}, \country{Vietnam}}}

\affil*[2]{\orgname{Vietnam National University}, \orgaddress{ \city{Ho Chi Minh City}, \country{Vietnam}}}

\abstract{Society needs to develop a system to detect hate and offense to build a healthy and safe environment. However, current research in this field still faces four major shortcomings, including deficient pre-processing techniques, indifference to data imbalance issues, modest performance models, and lacking practical applications. This paper focused on developing an intelligent system capable of addressing these shortcomings. Firstly, we proposed an efficient pre-processing technique to clean comments collected from Vietnamese social media. Secondly, a novel hate speech detection (HSD) model, which is the combination of a pre-trained PhoBERT model and a Text-CNN model, was proposed for solving tasks in Vietnamese. Thirdly, EDA techniques are applied to deal with imbalanced data to improve the performance of classification models. Besides, various experiments were conducted as baselines to compare and investigate the proposed model's performance against state-of-the-art methods. The experiment results show that the proposed PhoBERT-CNN model outperforms SOTA methods and achieves an F1-score of 67,46\% and 98,45\% on two benchmark datasets, ViHSD and HSD-VLSP, respectively. Finally, we also built a streaming HSD application to demonstrate the practicality of our proposed system.}


\maketitle

\section{Introduction}
\label{gioithieu}
Along with the advances in technology of the Fourth Industrial Revolution, the rapid rise of social networks has been astoundingly altering our daily life. In that situation, safety in cyberspace is an issue that directly affects the user's life, especially objects such as children or vulnerable people. According to Mohan et al. reports \cite{mohan2017impact}, the social media environment in which  many harmful contents such as hateful comments, fake news, contents that violate community standards influence not only on the large proportion of users but also on online moderators. 
Hate speech is typically described as any communication that disparages a person or group based on any attribute such as race, color, ethnicity, gender, sexual orientation, nationality, religion, or other trait. The following are some examples of hateful and offensive comments posted on Vietnamese social media\footnote{The several examples in this article are given to demonstrate the seriousness of the hate speech problem. They are based on actual online data and do not reflect the authors' opinions.}: cứ phải chửi cho mới chịu im :)))\textsubscript{you only shut your mouth until I swear at you :))))}; lũ chó đói\textsubscript{those hungry dogs}; hài vcl\textsubscript{so fucking funny}

However, censoring hate and offensive comments on social media faces many challenges because of their enormous volume and variety in both magnitude and topics. According to Suha Abu et al., research \cite{abu2018dental}, 293,000 posts are posted every 60 seconds on the billion-user social networking platform, Facebook, and over 510,000 comments are written there. Moreover, according to the prestigious statistics reporting site, Statista \cite{statista2018global}, Facebook had to remove more than 11.3 million pieces of offensive and hateful content globally in 2018. In 2019, YouTube also removed over 1.800 million comments that violated community standards, and these statistics have risen dramatically on both platforms. In 2020, Facebook must remove more than 81 million hateful and offensive posts, a seven times increase from 2018. While YouTube must remove over 4,800 million comments in 2020, this tripled the figures in 2019. 

The above results are the efforts of the two most prominent social networking platforms to stop hate and offense in their platforms globally. According to a report from the Wall Street Journal \cite{deepa2018facebook}, up to 2018, Facebook had to spend hundreds of millions of dollars on their content moderation teams. Besides, the major American technology information site - The Verge, reports that Google also has a team consisting of nearly 10,000 people responsible for the same task. However, this team still has many shortcomings. 

Firstly, despite the fact that Facebook is available in over 100 languages \cite{deepa2018facebook}, but only slightly more than half of them have dedicated content censorship teams. Meanwhile, Southeast Asia (including Vietnam) is a significant market for Facebook but still lacks human resources with advanced language skills. In addition, the social network environment in Vietnam is extremely toxic, according to a survey by Microsoft \cite{microsof2020tcivility}. 

Secondly, even though content censorship teams have been trained and informed about extremely hateful content, they need to deal with its consequences. Many of them have psychological disorders, and some even face Post-traumatic Stress Disorder (PTSD) \cite{keane1994posttraumatic}, which is prevalent after witnessing a heinous crime, and many are reported that they will never fully recover. 

Thirdly, the large corporations that own these social networks and research laboratories have allocated resources to develop systems using Artificial Intelligence to solve this dilemma, but they are still impractical. These systems utilize rich and quality data sources that originated from their social networking platforms. They possess leading-edge methods that can be applied across multilingual \cite{malmasi2017detecting}, making powerful systems that quickly classify hate and offensive content. However, it is difficult for these systems to recognize the content that lacks context, specialized culture native, and are slow to keep up with the continuously improving development in hateful content. Furthermore, these systems have not been studied extensively and deeply enough in Vietnamese due to Limitations remaining in building the systems which result in being unusable to solve this task. 

Therefore, our research contributes to HSD task in Vietnamese. Our research proposes a novel system that applies advanced natural language processing techniques to classify the hate and offensive comments on social networks towards a healthier, safer online space. Our research can handle issues as small as single comments to as vast as continuously processing enormous amounts of data in real-time. The main scientific contributions of our research are summarized as follows. 

\begin{enumerate}
    \item We implemented strict and efficient data pre-processing techniques to clean comments collected from social networking sites. These techniques enhance the data quality and significantly improve extracting information before training the model. 
    \item A novel HSD model is proposed to improve the performance of the task in Vietnamese. To this end, the various experiments were conducted with four state-of-the-art approaches: machine learning approaches, deep learning approaches, transfer learning approaches, and combined approach. Compared to our proposed PhoBERT-CNN model, these state-of-the-art approaches aid in developing the baseline models.
    \item We applied EDA techniques to the ViHSD dataset and the HSD-VLSP dataset to deal with imbalanced data and verify the effectiveness and necessity of data augmentation for the task of Vietnamese HSD.
    \item To demonstrate the usefulness of the proposed system, we built an application that continuously streams data from the massive data source of social media platforms to detect hate and offensive comments. 
\end{enumerate}

Our proposed system can be applied to online newspapers or websites having small comments quantity but require strict censorship, as well as huge social networks or forums. The system helps to improve the comprehensive censorship of offensive and hateful comments on cyberspace in Vietnam. We contribute to building a positive, civilized environment or satisfying the need to orient and protect vulnerable subjects such as the elderly and children. Moreover, the application is also a basis for agencies and organizations to evaluate and control management, psychological research, and education behaviors.

The rest of the paper is organized as follows. In Section \ref{fundamental}, we survey and describe an overview of the fundamentals of the HSD task and relevant studies. Our proposed approach is presented in detail through Section \ref{phuongphap}. Section \ref{ketqua} is our experimental results on the given datasets, contribution of each module in the PhoBERT-CNN model, and instructions and actual performance of the HSD application with streaming data. Finally, Section \ref{ketluanvahuongphattrien} is Conclusion and Future works. 

\section{Fundamental of Hate Speech Detection on Streaming Data}
\label{fundamental}
\subsection{A brief introduction to hate speech detection task}\label{related}
HSD and sentiment analysis have been inextricably linked \cite{schmidt2017survey}, and these tasks have  recently become popular topics in Natural Language Processing. In this section, we summarize the Vietnamese HSD task \cite{vu2020hsd,luu2021large}. This task aims to detect whether a comment on social media is HATE, OFFENSIVE, or CLEAN. Formally, the task is described as follows.

\textbf{Input}: Given Vietnamese comments on the social networks sites.

\textbf{Output}: One of three different labels is predicted by classifiers.

•	\textbf{Hate speech (HATE)} contains abusive language, which regularly bears the aim of insulting individuals or groups and might include hate speech, derogatory and offensive language. An item (post or comment) is identified as HATE if it (1) targets individuals or groups based on their characteristics; (2) demonstrates a transparent intention to incite harm or to market hatred; (3) may or might not use offensive or profane words.

•	\textbf{Offensive but not hate speech (OFFENSIVE)} is an item that could contain offensive words, but it does not target individuals.

•	\textbf{Neither offensive nor hate speech (CLEAN)} is a normal item. It is conversation, and expresses emotions normally, and it does not contain offensive language or hate speech is a normal item.

\begin{figure}[H]
    \centering
    \includegraphics[width=\linewidth]{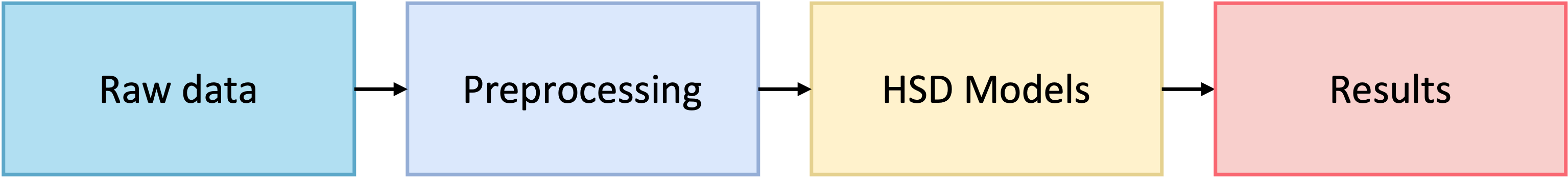}
    \caption{An overview of existing approach for hate speech detection.}
    \label{fig:overviewapproach}
\end{figure}

\subsection{Data pre-processing in HSD}
Data pre-processing techniques always play an essential role in data classification tasks from Vietnamese social networks in general and hate speech detection tasks in particular \cite{naseem2021survey}. Khang et al. \cite{nguyen2020exploiting} investigated the impact of pre-processing on datasets collected from Vietnamese social networks. According to the findings of this study, pre-processing has a significant impact on extracting information from data. Vietnamese comments on social media frequently contain characters and words associated with emotional undertones presenting in various ways, making it difficult to identify, differentiate, and extract information. Khang et al. \cite{nguyen2020exploiting} also succeeded in using initial data pre-processing to improve results by 4.66\%. This is a success compared to earlier work on the same datasets and evaluation metrics.

However, current studies on both the ViHSD and the HSD-VLSP datasets have not used modern and effective data pre-processing techniques to improve the performance of classification models. Only simple pre-processing techniques, such as word-segmenting texts into words, lower case text, removing or anonymizing sensitive information, and removing URLs and non-alphabetic characters, were utilized in previous studies. In our study, we inherit the advantages of previous studies and implement novel and specific pre-processing techniques to handle some of the most difficult challenges with social network data, such as SOTA Vietnamese word segmentation using VnCoreNLP \cite{vu2018vncorenlp}, De–teencode, and stopwords removal (see Subsection \ref{preprocess}). These techniques enhance the performance of models.

\subsection{Existing HSD models}
Some surveys on hate speech and machine learning for HSD give the research information on the current state of this field. They not only provide a structured overview of previous approaches \cite{fortuna2018survey} but also describe key sub-areas that have been explored to recognize these types of utterances automatically \cite{schmidt2017survey}. In addition to providing a survey and state-of-the-art natural language processing (NLP) technique that is used in the automatic detection of the hate speech on the Online social networks \cite{alrehili2019automatic}. This motivates other researchers, namely  Waseem et al. \cite{waseem2016hateful}, Chen et al. \cite{chen2018verbal}, and Davidson et al. \cite{davidson2017automated}, to apply the HSD system with the purpose of solving the real-life problem of hate speech on social network like Twitter. 

On the other hand, we conducted a survey on the related works serving the task of classifying comments on social networks in Vietnamese, especially the Vietnamese Hate Speech Detection task is still modest \cite{do2019hate,huu2019automated,huynh2020simple,luu2020comparison,luu2021large,nguyen2019vais,van1991nlp,vu2020hsd}. Specifically, the current studies revolve only based on two typical datasets by their outstanding high quality and large quantity of data points: ViHSD \cite{luu2021large} and HSD-VLSP \cite{vu2020hsd} datasets.

Methods for solving HSD problems are numerous, with machine learning models being the most fundamental. Support Vector Machine and Random Forest models applied in the study of Davidson et al.\cite{davidson2017automated} and Martins et al.\cite{martins2018hate} are the best approaches across their studies, and the results serve as the foundation for future development of other methods. In recent years, SOTA solutions with exceptional performance have emerged, such as the development of single models for multi-language such as BERT \cite{devlin2018bert}, RoBERTa \cite{liu2019roberta}, XLM-R \cite{conneau2019unsupervised}, and combinations to create more superior models such as those such as BERT-CNN \cite{safaya2020kuisail}, RoBERT-CNN\cite{liu2020hybrid}, XLMR-CNN \cite{saha2021hate}, which provide opportunities for HSD performance increase.

As inspired by the success of combining variant BERT with the CNN model \cite{safaya2020kuisail,liu2020hybrid,saha2021hate}, the PhoBERT-CNN combined model is implemented in this work to investigate its efficacy in the task of Vietnamese HSD. 

CNN is used instead of other typical deep neural networks such as LSTM \cite{hochreiter1997long}, Bi-LSTM \cite{schuster1997bidirectional}, and GRU \cite{chung2014empirical} since it is currently the most successful model for addressing short text classification tasks \cite{he2019using}. The convolution and pooling techniques of CNN aid in the extraction of the main concepts and keywords of the text as features, resulting in a significant improvement in the performance of the classification model. To address this limitation, the large-scale monolingual language model pre-trained for Vietnamese PhoBERT is the appropriate combination due to the fact that the PhoBERT has a duty on extracting features from sentences for the input of the Text-CNN model.

PhoBERT, the first large-scale monolingual pre-trained language model for Vietnamese, was introduced by Nguyen et al. in 2020 \cite{nguyen2020phobert}. PhoBERT was trained on about 20GB of data, including approximately 1 GB from the Vietnamese Wikipedia corpus and the rest of 19 GB from the Vietnamese news corpus. The architecture of PhoBERT is similar to the RoBERTa model developed by Liu et al. at Facebook \cite{liu2019roberta} (the model is optimized from the BERT model with a large amount of data training data up to 160GB and a 10x increase over BERT). Furthermore, when it comes to Vietnamese, PhoBERT has been demonstrated to perform and produce better results than the current best multilingual model \cite{nguyen2020phobert}, namely the XLM-R model \cite{conneau2019unsupervised}.

\subsection{Hate speech detection with streaming data}
Besides new datasets and methodologies, numerous applications and systems for real-time data processing are also introduced. Some typical projects can be mentioned, such as data streaming process of Nagarajan et al. \cite{nagarajan2019classifying}, and real-time tweets processing on Twitter, which uses Spark Streaming \cite{zaki2020real}. Some real-life applications have been introduced in the last decade, in 2015, Burnap et al. \cite{burnap2015cyber} successfully provided Application Program ming Interfaces (APIs) based on website services such as CrowdFlower or Amazon Mechanical Turk which can be integrated into a data pipeline in order to classify hate speech text. Then in 2018, Anagnostou et al. also presented a web application for actively reporting YouTube \cite{anagnostou2018hatebusters}. 

However, to the best of our knowledge, current research on HSD for Vietnamese is still at the theoretical analysis stage, and no practical application solutions have been introduced. Therefore, we need a highly scalable, reliable, and fault-tolerant data streaming engine for real-time HSD.

We also overcome the remaining restrictions on the ViHSD and HSD-VLSP datasets in previous studies \cite{luu2021large,vu2020hsd,do2019hate,nguyen2019vais,van1991nlp,huu2019automated,luu2020comparison,huynh2020simple} by proposing a two-phase data pre-processing techniques. Furthermore, we inherit the advantages of each study, such as the ability to conduct the experiment with deep learning, transfer learning models, and combined models. In particular, we combine a robust monolingual pre-trained model for Vietnamese, PhoBERT and a deep learning model, Text-CNN. At the end of the experiment, an HSD application with streaming data was deployed to demonstrate the effectiveness and contribution of our proposed system.

\section{Proposed Hate Speech Detection System for Social Media Streaming Data}
\label{phuongphap}
\subsection{Proposed System Architecture}
\begin{figure}[H]
    \centering
    \includegraphics[width=\linewidth]{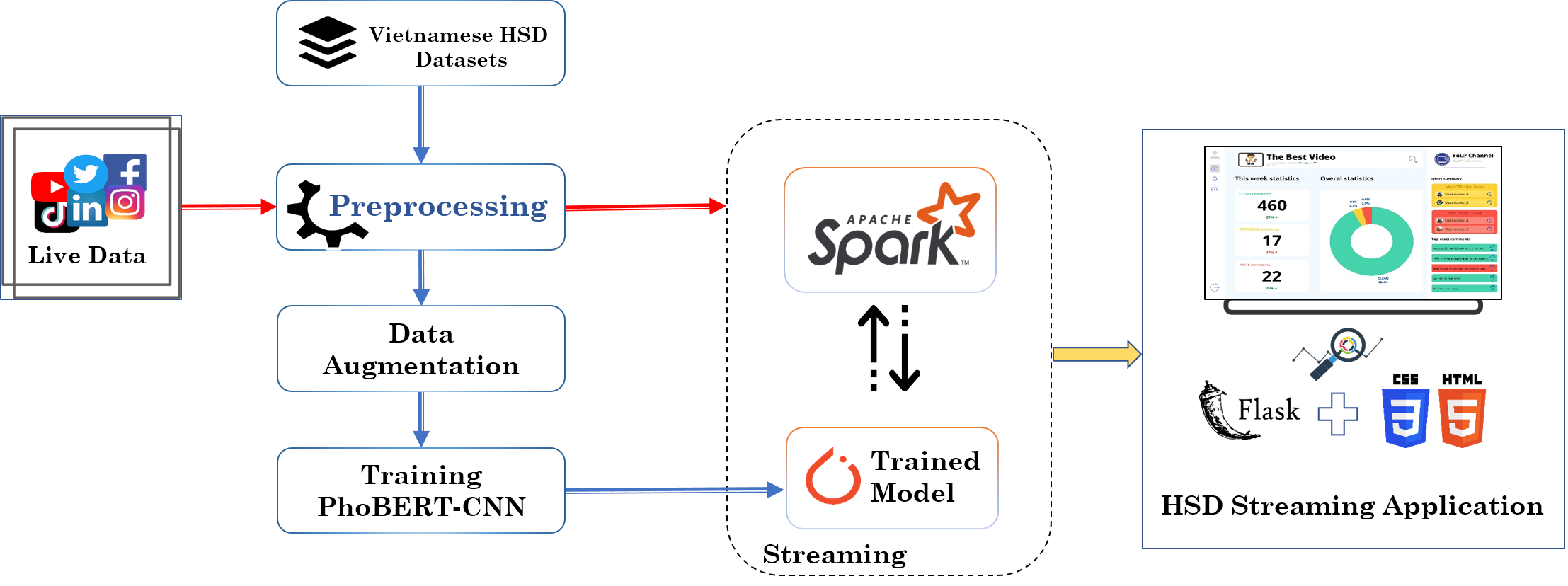}
    \caption{Our proposed approach for Vietnamese hate speech detection.}
    \label{fig:approach}
\end{figure}

This section proposes our efficient and straightforward approach for the Vietnamese hate speech detection task. We focus on perfecting the combined model, PhoBERT-CNN, for generating a best-performance model by fine-tuning techniques. Figure \ref{fig:approach} shows the overview of the system using four essential components: the pre-processing techniques, the data augumentation techniques, the core method PhoBERT-CNN, and the Hate speech detection application with streaming data.

\subsection{Data pre-processing}
\label{preprocess}

\begin{figure}[H]
    \centering
    \includegraphics[width=\linewidth]{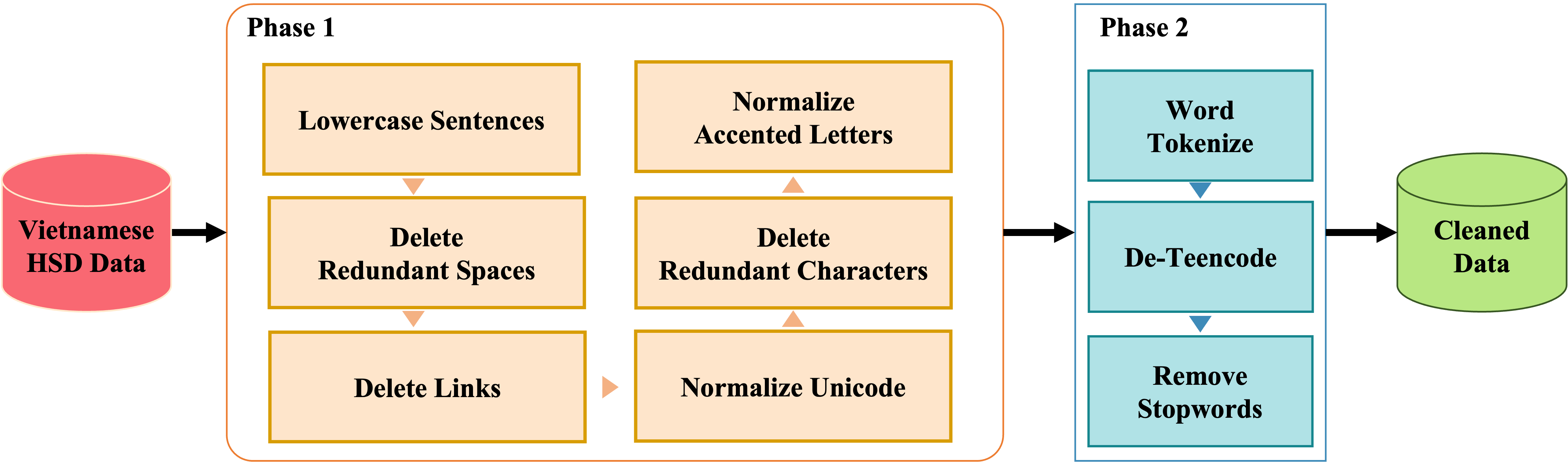}
    \caption{Data preparation steps.}
    \label{fig:data_preparation}
\end{figure}

We use two datasets: ViHSD \cite{luu2021large} and HSD-VLSP 
\cite{vu2020hsd} which contain 33,400 and 20,345 comments, respectively. Because the ViHSD and HSD-VLSP datasets are collected from social networking sites, they contain highly complex and diverse comments. Especially, abundant comments in both datasets contain, non-unicode standard characters, teen code, acronyms, and words with repeating characters. Therefore, we proceed to build a data pre-processing process to improve the quality of the datasets to extract valuable features before using them for training the classification models. Figure \ref{fig:data_preparation} depicts an overview of the two-phase data pre-processing procedure.

\subsubsection{Phase 1:}\label{phase1}\text{ \\}

\textbf{Lowercase Sentences}: All the characters of all the comments in the datasets are lowercase. We do this to avoid Python seeing two exact words as separate because of their capital letters.

\textbf{Delete Redundant Space}: Users on social media unwittingly or wittingly type multiple spaces on their comments. Therefore, we have decided to remove those redundant spaces.

\textbf{Delete Links}: We believe that website in comments do not affect the sentiment of the comment. As a result, we have also decided to remove them all.

\textbf{Normalize Unicode}: We also see a lot of Vietnamese words in the dataset that are the same, but Python detects them as separate because of their Unicode. The reason is there are many Unicode Transformation Formats (UTF) such as UTF-8, UTF-16, UTF-32 that are used widely, but our choice is normalizing to UTF-8.

\textbf{Delete Redundant Characters}: We remove redundant characters that the users intentionally make. 

\begin{figure}[H]
    \centering
    \includegraphics[width=1.\linewidth]{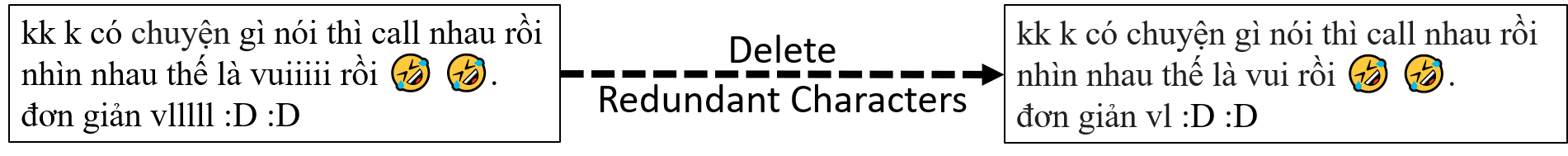}
    \caption{An example of a comment which generally means ``If have nothing to say, have a call to see each other is happy enough. So fucking simple'' and deleted redundant characters version of it.}
    \label{fig:redundant_char}
\end{figure}

Figure \ref{fig:redundant_char} above is an instance of the process of Deleting Redundant Characters. The word ``vuiiii'', which means happy, and the word ``vlllll'', which is an offensive teen code in Vietnamese but does not have a clear meaning in English, both after removing redundant characters, they become ``vui'' and ``vl''. However, the duplicated characters in the word ``kk'', which is also a teen code for the action of laughing - kaka, and ``call'' are ``k'' and ``l'', respectively, are not removed. Indeed, the word ``call'' will become ``cal'', which has no meaning, and the word ``kk'' will become ``k'', which is a different teen code meaning no in English.

\textbf{Normalize Accented Letters}: Because of the inconsistency in placing diacritical marks, we decide to normalize them in comments by followings rules:

\begin{itemize}
    \item If there is one vowel, the diacritical mark(s) will be on that vowel. For example: má (mom), lá (leaf), mê (love).
    \item If there are two vowels, the diacritical marks will be on the first one. For example: lóa (shinny), qùa (gift). If three vowels or two vowels follow with a consonant, the diacritical marks will be on the second vowel, for example: khuỷu (elbow), quán (store).
    \item ``ê'' and ``ơ'' are exceptional because the diacritical marks are always on them, for example: khuyển (dog), quở (reproachfully).
\end{itemize}

\begin{table}[ht]
\begin{center}
\begin{minipage}{\textwidth}
\centering
\caption{Number of changes on the datasets after the Phase 1.}
\resizebox{\textwidth}{!}{
\begin{tabular}{p{2cm}p{1.8cm}p{1.4cm}p{1.8cm}p{1.4cm}p{1.8cm}p{0.8cm}}
\hline
\multirow{2}{*}{\textbf{Datasets}} & \multirow{2}{*}{\textbf{Lowercase}} & \multicolumn{2}{c}{\textbf{Redundant}}                               & \multicolumn{2}{c}{\textbf{Inconsistent}}                                 & \multirow{2}{*}{\textbf{Link}}  \\ \cline{3-6}
                  &                            & \multicolumn{1}{l}{\textbf{Spaces}} & \multicolumn{1}{l}{\textbf{Characters}} & \multicolumn{1}{l}{\textbf{Unicode}} & \multicolumn{1}{l}{\textbf{Accented words}} &                        \\ \hline
\textbf{ViHSD}             & \multicolumn{1}{r}{28,540} & \multicolumn{1}{r}{488}                        & \multicolumn{1}{r}{2,127}                          & \multicolumn{1}{r}{753}                         & \multicolumn{1}{r}{620}                                & \multicolumn{1}{r}{21} \\
\textbf{HSD-VLSP}          & \multicolumn{1}{r}{0}      & \multicolumn{1}{r}{1}                          & \multicolumn{1}{r}{2,667}                          & \multicolumn{1}{r}{0}                           & \multicolumn{1}{r}{761}                                & \multicolumn{1}{r}{1}  \\ \hline
\end{tabular}
\label{tab:statistics}
}
\end{minipage}
\end{center}
\end{table}

All the above steps in Phase 1 are conducted in the same order. The output of this Phase 1 is fed directly to the next Phase 2.

\subsubsection{Phase 2:} \label{phase2}
\text{ \\} 
\quad \textbf{Word tokenize}: The input sentence is split into words or meaningful word phrases. In order to do this, we used Word Segmenter of VnCoreNLP \cite{vu2018vncorenlp} for the PhoBERT model and NLTK \cite{bird2006nltk} for the other models. Because the comments in both datasets, ViHSD \cite{luu2021large} and HSD-VLSP \cite{vu2020hsd}, are raw text data, word segmentation is required to prepare the data for PhoBERT model training \cite{nguyen2020phobert}. Moreover, PhoBERT used the VnCoreNLP RDRSegmenter \cite{vu2018vncorenlp} to pre-process the pre-training data \cite{nguyen2020phobert} (including Vietnamese word and sentence segmentation), it is recommended that the same word segmenter must be used for PhoBERT downstream applications in relation to the input raw texts. On the other hand, the other models could learn from text data at the token level without requiring word segmentation, as the PhoBERT model does. As a result, we decided to tokenize the pre-training data using NLTK \cite{bird2006nltk}.

\textbf{De–teencode}: In social networks, people usually spend a significant amount of their time to chit chat and also often use the short form of words to type faster. Some are used to trick the systems when they are swearing. Moreover, those abbreviations also have their name in Vietnam, teen codes. As a result, to help our models better understand the input sentences, we had to map those teen codes into their original words. Furthermore, the process of mapping teen codes, we named it De–teencode, and the following Table \ref{tab:example_of_teencode} shows some instances of them.

\begin{table}[ht]
\begin{center}
\begin{minipage}{\textwidth}
\centering
\caption{Examples of teencodes and their expansions.}
\label{tab:example_of_teencode}
\begin{tabular}{llll}
\hline
\multirow{2}{*}{\textbf{No.}} & \multirow{2}{*}{\textbf{Teencode}} & \multicolumn{2}{l}{\textbf{De-Teencode}} \\ \cline{3-4} 
                              &                                    & \textbf{Vietnamese sentences} & \textbf{English meanings} \\ \hline
\multicolumn{1}{r}{1}                             & đc đấy                             & được đấy             & nice              \\
\multicolumn{1}{r}{2}                             & ko                                 & không                & no                \\
\multicolumn{1}{r}{3}                             & cc                                 & con c*c              & d*ck              \\ \hline
\end{tabular}
\end{minipage}
\end{center}
\end{table}

\textbf{Remove stopwords}: We also removed stopwords from the comments because of their meaninglessness. In our experiments, we used the Vietnamese stopword dictionary \cite{le2017stop} to remove stop words in the sentence.

In Phase 2, the data are tokenized, De-teencode, and removed stopwords. Phase is in that order since the output of Word tokenizers is a list of words, word phrases, and characters that are separate from the others by space. Those characters then are checked if they are teen code and will be De–teencode in the next step. Therefore, the De–teencode step following after the Word tokenize step is a wise decision. Finally, after the De–teencode step, we remove all stopwords, and the reason we remove stopwords after De–teencode step is that those teencodes possibly are also stopwords. Table \ref{tab:TeenStop_count} presents basic statistics about teencode and stopword words in two datasets, ViHSD and HSD-VLSP. From the statistics, we can observe that the relatively high proportions of the terms teencode and stopword in the two data sets indicate the use of several acronyms and abbreviations by social media users.

\begin{table}[h]
\begin{center}
\begin{minipage}{\textwidth}
\caption{The number of teencodes and stopwords of the datasets.}
\label{tab:TeenStop_count}
\begin{tabular}{llllll}
\hline
\multirow{2}{*}{\textbf{Datasets}} & \multicolumn{2}{c}{\textbf{Teencodes}}   & \multicolumn{2}{c}{\textbf{Stopwords}}   & \multicolumn{1}{c}{\multirow{2}{*}{\textbf{\#Words}}} \\ \cline{2-5}
                  & \textbf{Frequency} & \textbf{Percentage} & \textbf{Frequency} & \textbf{Percentage} & \multicolumn{1}{c}{}                                          \\ \hline
\textbf{ViHSD}    & \multicolumn{1}{r}{15,344}             & \multicolumn{1}{r}{4.00\%}              & \multicolumn{1}{r}{153,330}            & \multicolumn{1}{r}{40.01\%}             & \multicolumn{1}{r}{383,270}                                                       \\
\textbf{HSD-VLSP} & \multicolumn{1}{r}{13,757}             & \multicolumn{1}{r}{3.24\%}              & \multicolumn{1}{r}{127,531}            & \multicolumn{1}{r}{30.01\%}             & \multicolumn{1}{r}{424,301}                                                       \\ \hline
\end{tabular}
\end{minipage}
\end{center}
\end{table}

\subsection{Dealing with imbalanced data}\label{dealing}
More and more effective solutions for the HSD task based on pre-trained language models have recently been proposed. The development of various large-scale pre-trained language models is driving up demand for high-quality, large-scale data sources. One of the most important priorities in this context is the distribution of data samples, particularly the data balance, which has a significant impact on model evaluation performance \cite{luu-etal-2020-empirical}. However, there is a significant difference in the number of CLEAN comments compared to OFFICIAL and HATE comments in many real-world situations, particularly in the two experimental datasets, ViHSD and HSD-VLSP, resulting in a skewed sample distribution. Learning algorithms will be biased toward the majority group due to the above problem, making this a challenging and exciting task to deal with in this paper. Meanwhile, the minority classes, such as HATE and OFFENSIVE, are typically more beneficial in terms of information mining, as they contain crucial information for the task of HSD despite its quite rare. To address this challenge, we focused on developing an intelligent system that used efficient data preprocessing techniques, suitable data augmentation, and a robust classification model to overcome bias. This is known as learning from imbalanced data \cite{10.5555/1293951.1293954}.

As a result, inspired by the study of Wei et al. \cite{wei-zou-2019-eda}, we intend to apply EDA techniques with four operations: synonym substitution, random insertion, random swap, and random deletion to deal with imbalanced data. These techniques are applied to the ViHSD training set and the HSD-VLSP dataset in this paper with the percentage of replacement word in the sentence ($\alpha$) equal to 0.15.

\begin{table}[h]
\centering
\caption{The ViHSD training set and HSD-VLSP dataset overview statistics before and after data augmentation.}
\label{tab:statistics_aug}
\resizebox{\textwidth}{!}{%
\begin{tabular}{llrrrrrr}
\hline
\multicolumn{1}{c}{\multirow{2}{*}{\textbf{Dataset}}} &
  \multicolumn{1}{c}{\multirow{2}{*}{\textbf{Label}}} &
  \multicolumn{3}{c}{\textbf{Original dataset}} &
  \multicolumn{3}{c}{\textbf{Augmented dataset}} \\ \cline{3-8} 
\multicolumn{1}{c}{} &
  \multicolumn{1}{c}{} &
  \multicolumn{1}{c}{\textbf{\begin{tabular}[c]{@{}c@{}}Num. \\ Comments\end{tabular}}} &
  \multicolumn{1}{c}{\textbf{\begin{tabular}[c]{@{}c@{}}Avg. \\ Word length\end{tabular}}} &
  \multicolumn{1}{c}{\textbf{\begin{tabular}[c]{@{}c@{}}Vocab. \\ size\end{tabular}}} &
  \multicolumn{1}{c}{\textbf{\begin{tabular}[c]{@{}c@{}}Num. \\ Comments\end{tabular}}} &
  \multicolumn{1}{c}{\textbf{\begin{tabular}[c]{@{}c@{}}Avg. \\ Word length\end{tabular}}} &
  \multicolumn{1}{c}{\textbf{\begin{tabular}[c]{@{}c@{}}Vocab. \\ size\end{tabular}}} \\ \hline
\multirow{3}{*}{\textbf{\begin{tabular}[c]{@{}l@{}}ViHSD \\ training set\end{tabular}}} &
  CLEAN & 19,886 & 6.55 & 130,238 & 19,886 & 6.55 & 130,238 \\&
  OFFENSIVE & 1,606 & 7.24 & 11,624 & 10,147 & 7.57 & 76,802 \\&
  HATE & 2,556 & 12.08 & 30,883 & 16,849 & 11.64 & 196,086 \\ \hline
\multirow{3}{*}{\textbf{HSD-VLSP}} &
  CLEAN & 18,614 & 14.85 & 276,557 & 18,614 & 14.85 & 276,557 \\&
  OFFENSIVE & 1,022 & 8.87 & 9,063 & 8,461 & 8.05 & 68,093 \\&
  HATE & 709 & 14.23 & 10,087 & 6,392 & 13.41 & 85,713 \\ \hline
\end{tabular}%
}
\end{table}

We first applied the EDA techniques on the ViHSD training set and the entire original HSD-VLSP dataset to enhance the data on HATE and OFFENSIVE labels. Table \ref{tab:statistics_aug} describes the information about the HSD-VLSP dataset after making data augmentation.

\begin{figure}[H]
\centering
    \begin{subfigure}[ht]{0.49\textwidth}
    \centering
    \includegraphics[width=\linewidth]{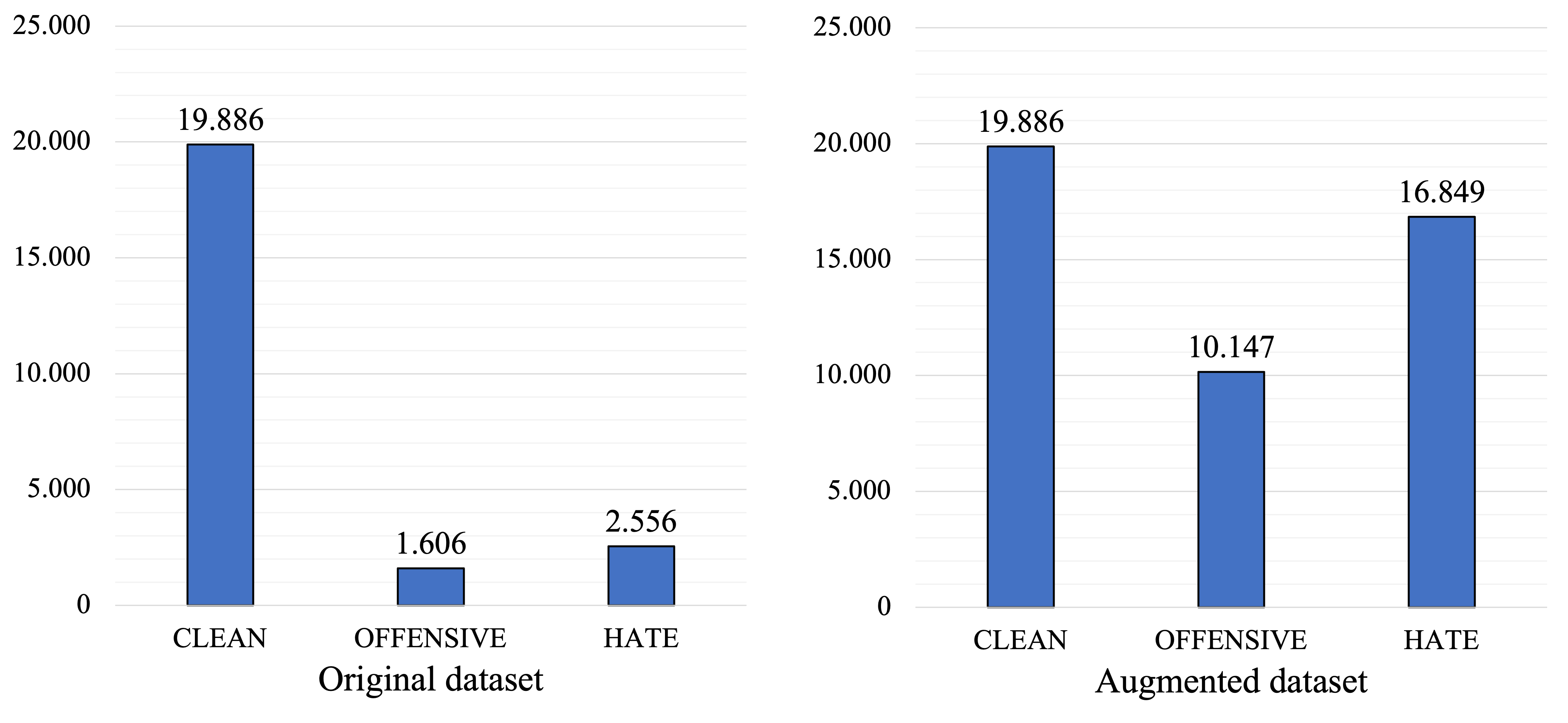}
    \caption{ViHSD training set.}
    \label{fig:vihsd_aug}
    \end{subfigure}
\hfill
    \begin{subfigure}[ht]{0.49\textwidth}
    \centering
    \includegraphics[width=\linewidth]{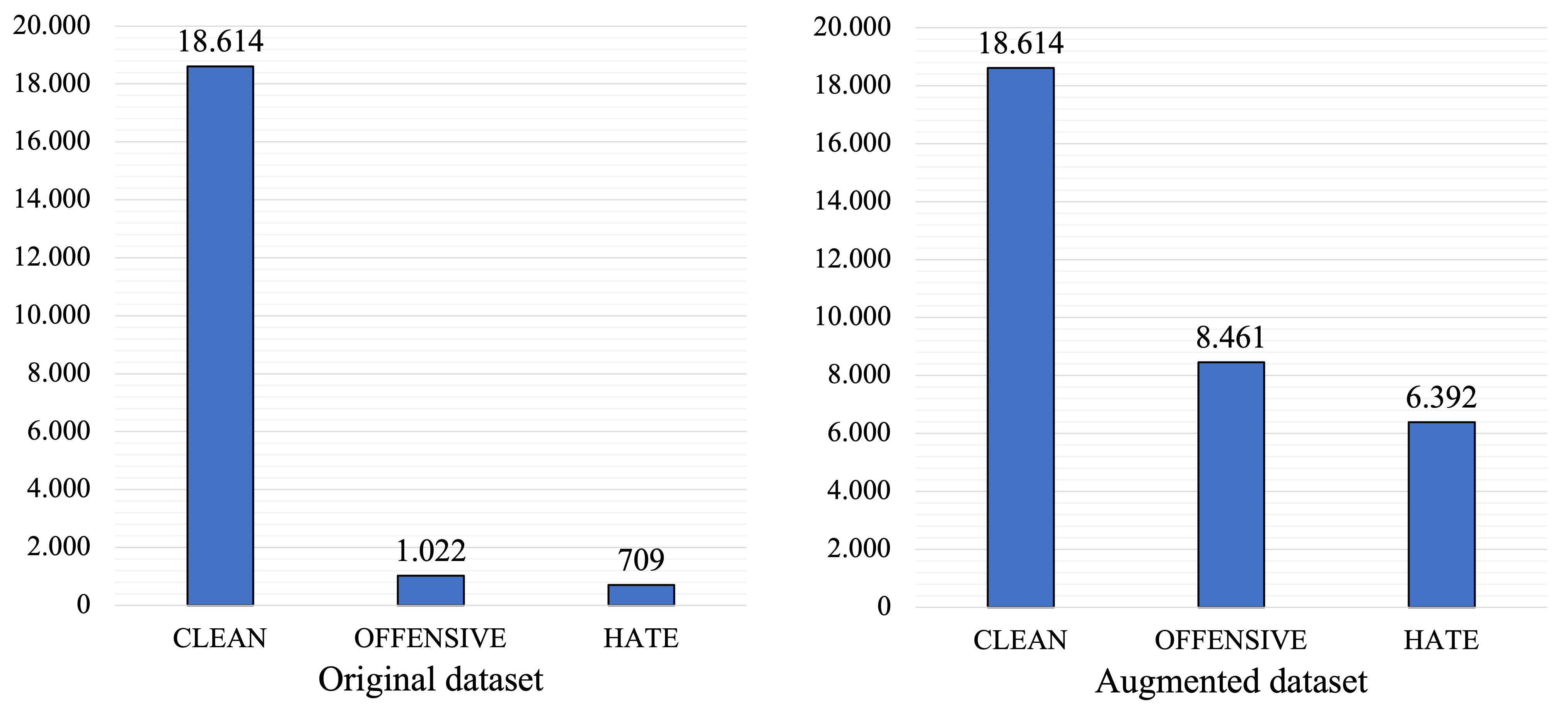}
    \caption{HSD-VLSP dataset.}
    \label{fig:hsd_aug}
    \end{subfigure}
    \caption{The labels distribution in the ViHSD training set and the HSD-VLSP dataset before and after augmentation.}
    \label{fig:vidata_aug}
\end{figure}

Table \ref{tab:statistics_aug} shows that after applying EDA techniques, the number of data and vocabulary size on the HATE and OFFENSIVE labels grew dramatically. Figure \ref{fig:vidata_aug} depicts the label distribution in the ViHSD training set and the HSD-VLSP dataset before and after augmentation. Compared with the original dataset, the data on augmented dataset are well-distributed after utilizing EDA methods.

\subsection{Proposed PhoBERT-CNN model for HSD}
\label{phobertcnn}
\begin{figure}[ht]
    \centering
    \includegraphics[width=\linewidth]{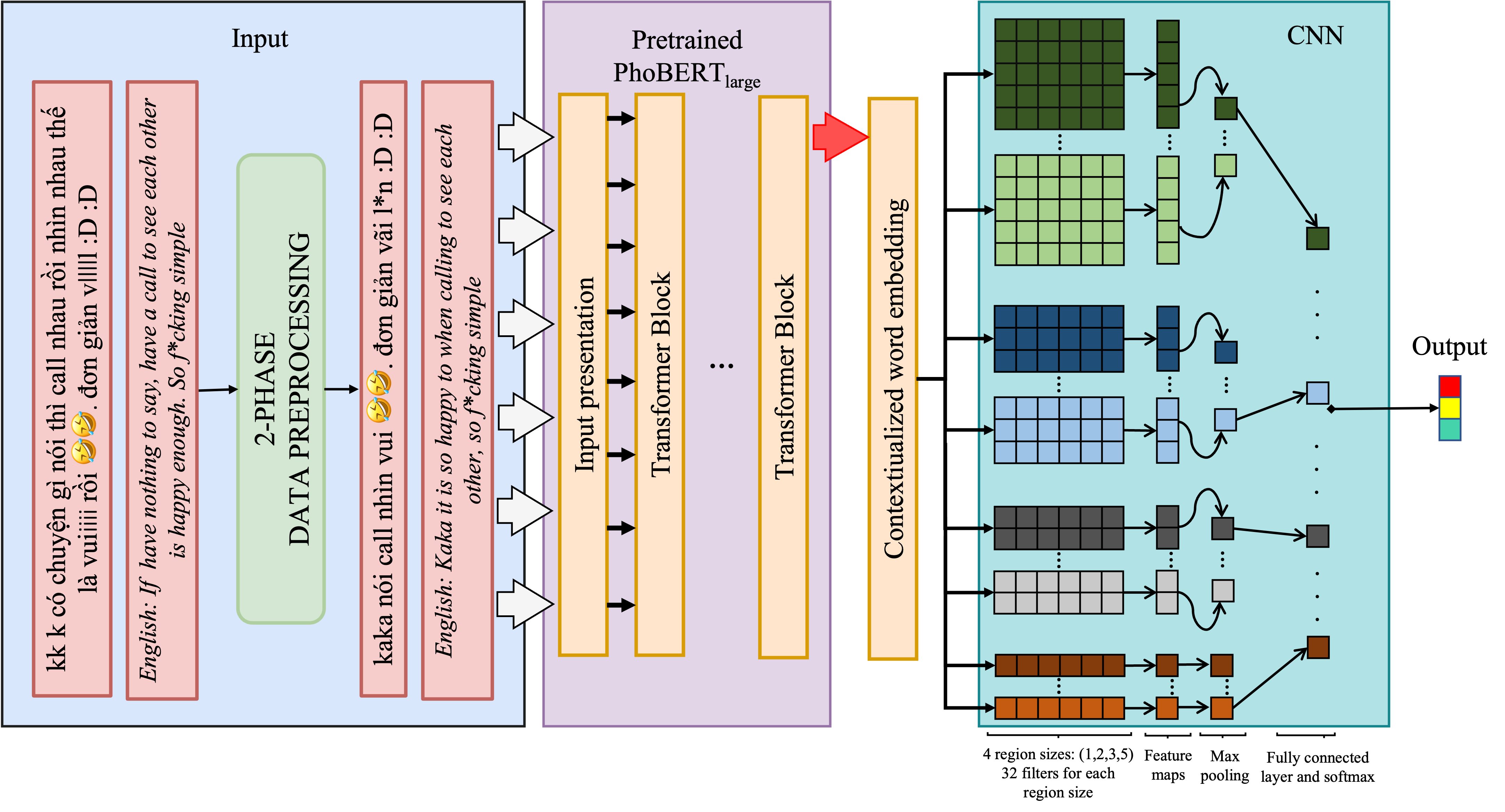}
    \caption{An overview of our Vietnamese HSD system using PhoBERT-CNN.}
    \label{fig:architecture}
\end{figure}

Variant BERT and CNN combined models have recently been widely used to classify short text collected from social networks, particularly to classify hate and offensive comments that achieve promising results \cite{safaya2020kuisail,liu2020hybrid,saha2021hate}. As a result, in this work, the PhoBERT and CNN combined model is deployed to evaluate their efficacy in classifying hate and offensive comments for Vietnamese. The combined PhoBERT-CNN model is expected to significantly improve the classification performance thanks to the resonance mechanism of the two single models, reducing errors between the predicted labels and the actual labels. Both of the two single models, PhoBERT and Text-CNN, outperformed other models on the tasks of classifying Vietnamese text, particularly on the ViHSD and HSD-VLSP datasets \cite{luu2021large,vu2020hsd,do2019hate,nguyen2019vais,van1991nlp,huu2019automated,luu2020comparison,huynh2020simple}. Figure \ref{fig:architecture} presents the architecture of our approach for Vietnamese HSD.

Firstly, we present the architecture of PhoBERT \cite{nguyen2020phobert} and how the PhoBERT model is adapted to perform as a word embedding layer to extract information from data. The PhoBERT model was chosen because it outperforms previous monolingual and multilingual pre-trained language model approaches, achieving new state-of-the-art performance on four downstream Vietnamese NLP tasks, including Vietnamese Hate Speech Detection \cite{luu2021large,vu2020hsd,do2019hate,nguyen2019vais,van1991nlp,huu2019automated,luu2020comparison,huynh2020simple}. The architecture of the PhoBERT model is a multilayer architecture comprised of multiple layers of Bidirectional Transformer encoder. It takes the representation of a text sentence composed of a string by the contextualized words as input. The input representation of the PhoBERT model is built by summing those tokens with the segment vectors and the corresponding positions of the words in the sequence.

\begin{figure}[ht]
    \centering
    \includegraphics[width=\linewidth]{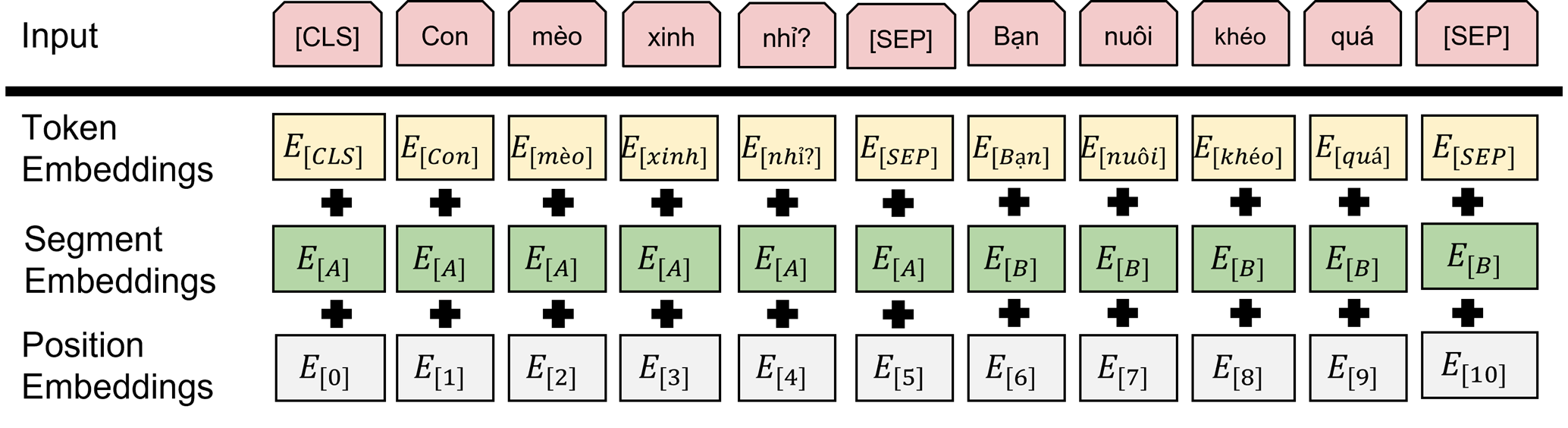}
    \caption{The process of representing the input of PhoBERT model.}
    \label{fig:inputPhoERT}
\end{figure}

\begin{itemize}
    \item We use Positional Embeddings with a maximum sentence length of 20 words.
    \item  The first word for each string defaults to the special word \texttt{[CLS]}. The output of the final hidden state corresponding word \texttt{[CLS]} will be used to represent the whole sentence in the classification.
    \item When a string only comprises a single sentence, the embedding segment can be applied directly to that sentence.
    \item In case the string contains more than two-sentence, we distinguish sentences in two steps: separate sentences by a special word called \texttt{[SEP]} and we add independent segment embedding for each statement.
\end{itemize}

Next, the Fully-connected layer at the end of the pre-trained PhoBERT model is replaced with a CNN network architecture \cite{kim2014convolutional}. Because CNN is currently the most successful model for solving short text classification tasks \cite{he2019using}, it is utilized instead of other typical deep neural networks such as LSTM, Bi-LSTM, and GRU \cite{pham2020pgsg,li2019exploiting,yi2019pre}.

In this research, we construct the layers of the CNN model as follows:
\begin{itemize}
    \item \textbf{INPUT}: The input layer is used to initialize the input object from a matrix of vectors. The input and output objects of this layer have dimensions equal to the number of dimensions of the vector matrices.
    \item \textbf{CONV1D}: We build four convolution layers using the Conv1D layer to extract features from a matrix of vectors. For each Conv1D, we all use a filter of a particular size, set kernel size values, and use the ReLU activation function to enhance convergence.
    \item \textbf{POOLING}: We perform maximal pooling (Max Pooling) using the MaxPool1D function. The output of this layer is a matrix of features that have been reduced in size but still retain the features extracted from the previous Conv1D layer. 
    \item \textbf{DROPOUT}: Before constructing the dropout layer, we use the torch.cat() function to concatenate the feature matrices that have been Pooled from the previous layer. Then, we set dropout values to remove random notes from each hidden layer.
    \item \textbf{FC}: In this layer, we construct a loss function and an optimized function to connect input data from the dropout layer to the classification classes. We utilize the Adam optimization algorithm and the CrossEntropy loss function (Equation \ref{eqn:loss}) for optimization: 
    \begin{align}\label{eqn:loss}
        -\sum_{c=1}^My_{o,c}\log(p_{o,c})
    \end{align}
    where $\textbf{M}$ is the number of classes (CLEAN, OFFENSIVE, HATE), $\textbf{log}$ is the natural log, $\textbf{y}$ is binary indicator (0 or 1) if class label $\textbf{c}$ is the correct classification for observation $\textbf{o}$, $\textbf{p}$ is predicted probability observation $\textbf{o}$ is of class $\textbf{c}$.
\end{itemize}

The convolution and pooling techniques of CNN aids in the extraction of the main concepts and keywords of the text as features, resulting in a significant improvement in the performance of the classification model. However, the CNN network has a significant limitation which is not suitable for sequence-level text \cite{kim2014convolutional,he2019using}. To address this limitation, the large-scale monolingual language model pre-trained for Vietnamese PhoBERT is the appropriate combination due to the fact that the PhoBERT has a duty on extracting features from sentences for the input of the Text-CNN model. Following that, a contextualized word embedding of comments from PhoBERT is fed into the Text-CNN model to get the feature maps. Finally, the prediction labels are given through a softmax layer.

The combined PhoBERT-CNN model will significantly improve the classification performance thanks to the resonance mechanism of the two single models, reducing errors between the predicted labels and the actual labels (see Subsection \ref{ablationanalysis}).

\subsection{Integrating HSD Model to Streaming Processing System}
\label{streaming}
One of the top-priority requirements is an advanced method for handling hate and offensive comments in big data environments such as social networks. This helps social networks in general and social networks in Vietnam, identify hate and offensive comments better, and also reduce the workload of moderators. Furthermore, the media agencies need an automatic moderator tool to more precisely monitor the comments that are permitted to be displayed.

We built an application for comment analysis capable of continuously collecting content from social networking sites to analyze comment nuances to meet these needs. These platforms will provide an API to send processing comments requests to the classification application.

\subsubsection{Streaming Data Ingestion}
Data streaming is the process of collecting data continuously in real-time from multiple data sources that are often put into stream processing applications to get essential insights.

Data streaming is critical for dealing with enormous amounts of live data. Such data can come from various sources, especially social networking sites like Facebook, Youtube, and Twitter.

Several real-time data streaming approaches are available, including Apache Kafka, Spark Streaming, and Apache Flume. In this paper, we will implement data streaming using Spark Streaming \cite{zaharia2016apache}.

\begin{figure}[ht]
    \centering
    \includegraphics[width=\linewidth]{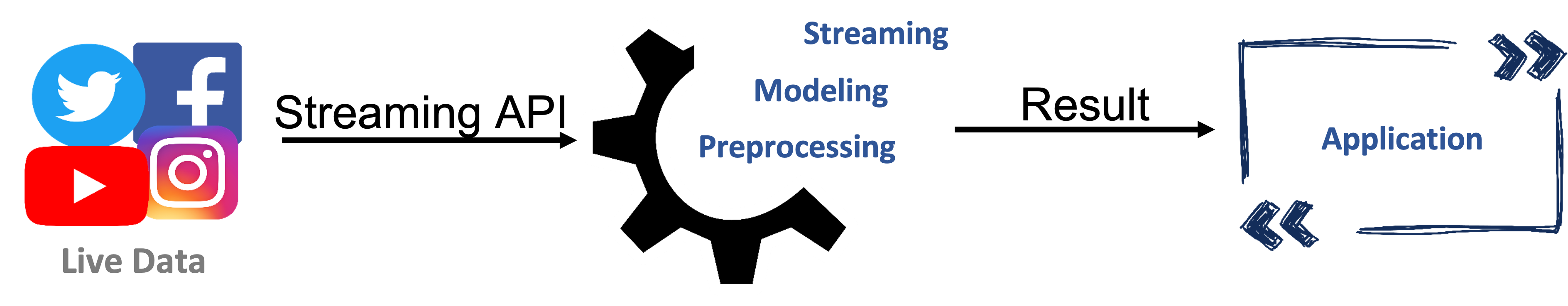}
    \caption{An overview of our Vietnamese HSD streaming system.}
    \label{fig:streaming}
\end{figure}

This section presents an end-to-end architecture on how to stream data from social media platforms, clean them, and apply the combined PhoBERT-CNN model to detect the hate or offense of each data. Figure \ref{fig:streaming} depicts an overview of the Vietnamese hate speech detection with a streaming system based on the PhoBERT-CNN model.

\textbf{Input data:} Live data collected through streaming API. 

\textbf{Main model:} Data pre-processing and hate speech detection on the collected data. 

\textbf{Output:} A parquet file with all the data and their hate speech prediction. 

\subsubsection{Instructions Streaming Processing for HSD}
We successfully constructed a system capable of handling large amounts of data in real-time from social networking platforms, especially from YouTube comments, by conducting surveys and experiments on processing streaming data \cite{nagarajan2019classifying,zaki2020real,anagnostou2018hatebusters,burnap2015cyber}. This section presents the instructions of the system in practice.

\paragraph{Part 1 - Send comments from the Youtube Data API:} 

In this part, we authenticate and connect to the Youtube Data API using our developer credentials. First, we need to set up the essential information to log in and utilize the Youtube Data API, such as DEVELOPER\_KEY, YOUTUBE\_API\_SERVICE\_NAME, YOUTUBE\_API\_VERSION. Next, in the QUERY section, we change the URL to query the video that needs to be processed. In addition, parameters such as textFormat to set the format of return comments, maxResults to set the number of comments are limited to each session. We also build a TCP socket between the Youtube Data API and Spark \cite{zaharia2016apache}, which waits for the Spark Streaming call and delivers data.

\paragraph{Part 2 - Data pre-processing and hate speech detection:}

Comment data collected and stored through Youtube Data API will be transmitted to the primary system to perform pre-processing according to the process described in Section \ref{preprocess}. 

After pre-processing, the normalized and high-quality data are used to predict their labels using the combined PhoBERT-CNN model. We use SparkSQL to query and visualize how Spark Streaming \cite{zaharia2016apache} organizes and presents predictions relative to comments. The prediction results in 0.0, 1.0, and 2.0 are corresponding to the CLEAN, OFFENSIVE, and HATE labels , which indicate the polarity of nuance provided by the comments. 

Finally, the query results are converted to DataFrame format so that the administrator can observe and monitor it more efficiently and are stored in a parquet file. These findings will be used to assist administrators in deciding whether to delete comments containing hate or offensive content or not.

\begin{figure}[H]
    \centering
    \includegraphics[width=0.8\linewidth]{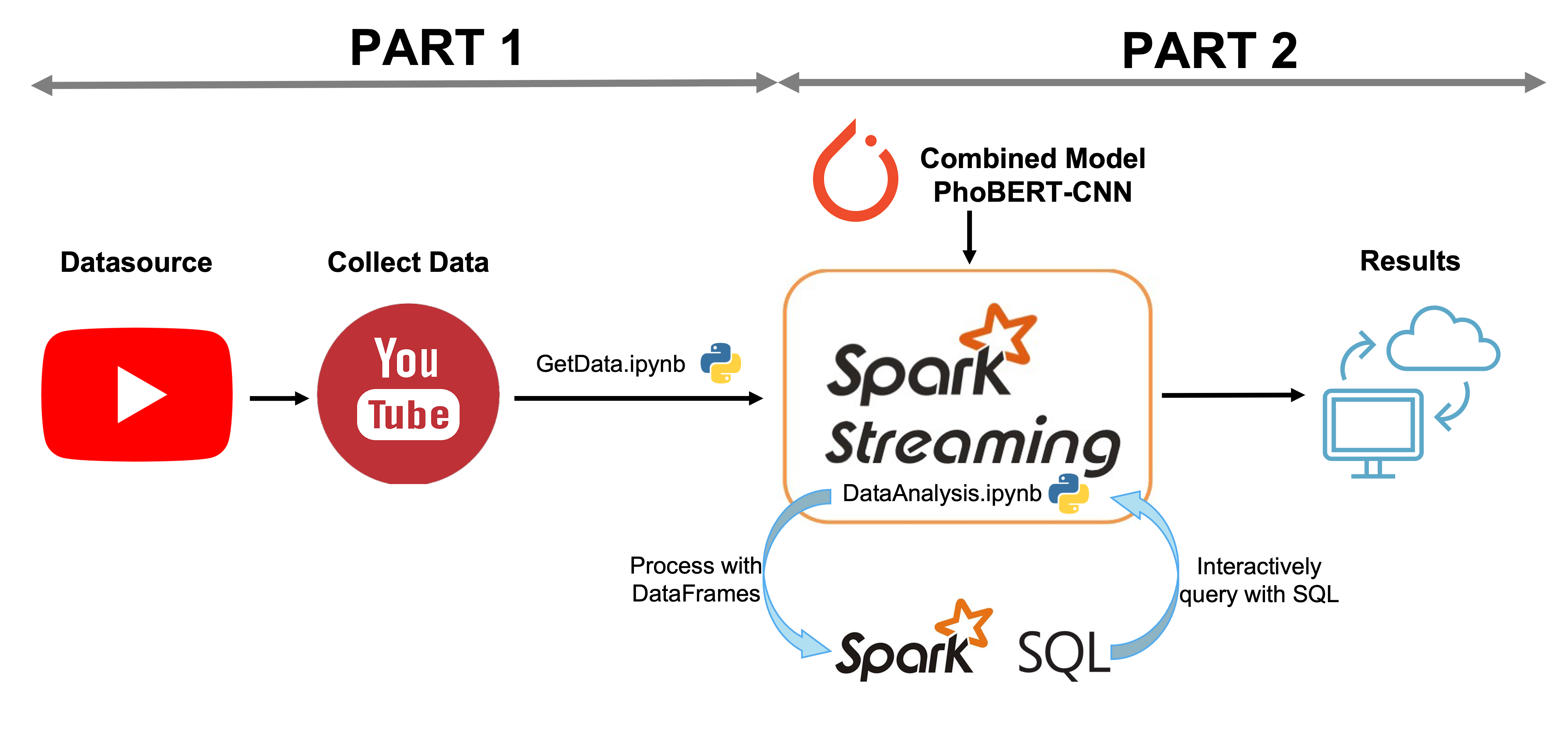}
    \caption{The end-to-end architecture of our system using Spark Streaming.}
    \label{fig:stream}
\end{figure}

\subsubsection{Reproducibility of the Proposed HSD System}
We also provide a solution constructed in PyTorch for reproducibility to make it easy to apply our novel proposed system for Vietnamese HSD. The code for our trained models and streaming application is publicly available for research purposes at \url{https://github.com/khanhtran0412/ViSoMeCens}.

\section{Experiments and Results}
\label{ketqua}

\subsection{Experimental Procedure}
This section provides an outline of how we conducted experiments in order to propose a new and efficient HSD system for Vietnamese. Firstly, the hate speech data is collected from social media, especially the two datasets ViHSD and HSD-VLSP will be cleaned using the data pre-processing process as described in Subsection \ref{preprocess}. Accordingly, the data after cleaning and quality assurance is used to train our baselines and proposed model. With each HSD model conducted, we fine-tune the hyperparameters to find the optimal hyperparameters and help improve the performance of the model. We also apply data augmentation techniques to the ViHSD training set and HSD-VLSP dataset to address the challenging problem of data imbalance. Next, we evaluated the performance of the models we conducted using the F1-macro and Accuracy metrics. The model evaluation results are described in detail in Subsection \ref{results}. Based on the achieved results in the F1-score, we choose the model with the best performance to conduct an error analysis on the wrong predictions discovered in our system.

\begin{figure}[H]
    \centering
    \includegraphics[width=\linewidth]{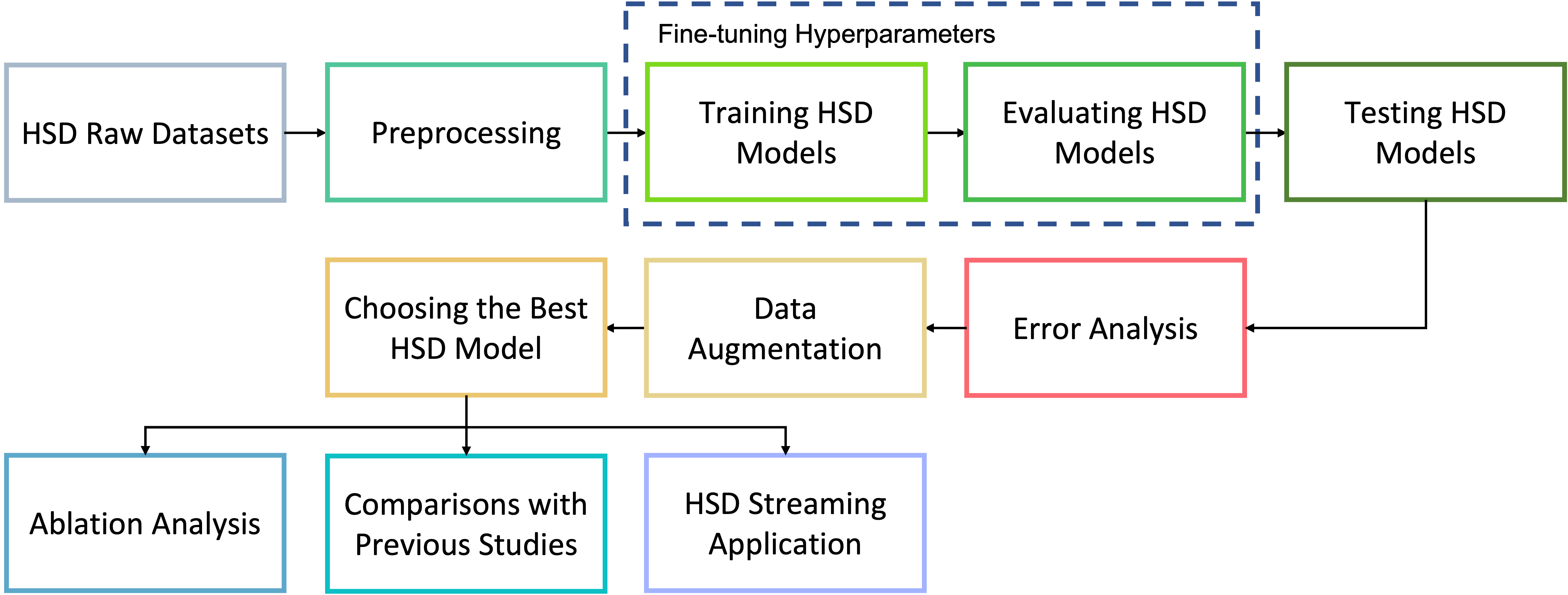}
    \caption{Overview of the experimental procedure for Vietnamese HSD.}
    \label{fig:exp_pro}
\end{figure}

In addition, comparisons with previous studies were made to assess the development of our study. Besides, an ablation analysis was carried out to investigate the effectiveness and contribution of our proposed PhoBERT-CNN approach. At the end of the experiment, a pilot experiment was performed to evaluate the actual performance of the proposed system when dealing with hate speech detection from data streaming in which our system was deployed to a social network, especially Youtube. Figure \ref{fig:exp_pro} presents an overview of the experimental procedure in this paper.

\subsection{Baseline models for HSD performance comparison} \label{baselinehsd}

\subsubsection{Machine learning approach}

\quad \quad \textbf{Multinomial Naive Bayes:} This algorithm predicts and classifies data based on observable data and statistics, using the Bayes theorem of probability theory. \cite {rish2001empirical,kim2002effective,liu2014combining}. 

Multinomial Naive Bayes is a supervised learning algorithm that is commonly used in machine learning because it is relatively easy to train and achieve high performance.

\quad \textbf{Logistic Regression:} Logistic regression is a binary classification algorithm, it is a simple, well-known, and important method in the discipline of machine learning. In addition, this algorithm is also used in a machine learning application to classify incoming data based on previous data.

By analyzing the relationship between all of the existing independent variables, the Logistic Regression model predicts a dependent data variable. In natural language processing, this method requires manual features extracted from data for text classification \cite {genkin2007large,hosmer2013applied,davidson2017automated,luu2020comparison}.

\quad \textbf{Decision Tree:} Decision tree is a supervised learning algorithm, it is the most powerful and popular method for classification \cite{pranckevivcius2017comparison,ikonomakis2005text,burnap2016us}. Decision tree algorithm is also known as a structure tree, where each node represents a test on an attribute, each branch is an outcome of the test, and each leaf node is a class label.

This approach used basic rules from training data to predict the class or value of the target variable. Specifically, the record started from the root of the tree and compared the attribute with the node attribute at each branch in the decision tree before predicting a final class label in the leaf node.

\quad \textbf{Random Forest:} Random Forest is a Supervised learning method used to solve classification and regression tasks. It is built on multiple sets of Decision Tree and the output of this algorithm is based on the aggregate decision on the decision trees it generates with the voting method.

However, we cannot understand how this algorithm works due to the complicated structure of this model and this is one of the Black Box methods. \cite{liaw2002classification,islam2019semantics,davidson2017automated,badjatiya2017deep}.

\subsubsection{Deep learning approach}
\quad \textbf{Convolutional Neural Network (Text-CNN)}: A convolutional neural network (CNN) is a multistage Neural network architecture developed for classification \cite{kim2014convolutional}. By using convolutional layers, it can detect combination features. Our experiments employ four convolutional layers with 32 filters for each layer. Finally, the softmax function uses the result to predict the label for the text.
    
\textbf{Bidirectional Long Short-Term Memory (Bi-LSTM)}: The Bi-LSTM \cite{schuster1997bidirectional} is a famous variant of RNN \cite{medsker1999recurrent}. The Bidirectional Long Short Term Memory can be trained using all available input information within the past and way forward for a selected timeframe. This method is robust in classification problems, and most of its achieved high-performance classification results. Therefore, we plan to choose it to compare with other classification models during this task.

\subsubsection{Transfer learning approach}
The transfer learning model has attracted increasing attention from NLP researchers worldwide for its outstanding performance. One of the SOTA language models as BERT, which stands for Bidirectional Encoder representations from transformers, is published by Devlin et al. \cite{devlin2018bert}. BERT and its variations (BERTology) \cite{tenney2019bert,michel2019sixteen,rogers2020primer} such as RoBERTa \cite{liu2019roberta}, XLM-R \cite{conneau2019unsupervised}, PhoBERT \cite{nguyen2020phobert} have almost dominated and asserted their strength on natural language processing tasks, even for Vietnamese hate speech detection tasks \cite{luu2021large,luu2020comparison,huynh2020simple}. For the aforementioned reasons, we decided to use BERT and its variants to find the optimal solution with good performance and contribute to the successful construction of our proposed solution.

\textbf{BERT \cite{devlin2018bert}}: is a contextualized word representation model pre-trained using bidirectional transformers and based on a masked language model. In this work, we use the train set to fine-tune the pre-trained BERT model before classifying comments or posts from websites or social networks.

\textbf{Robustly optimized BERT approach (RoBERTa) \cite{liu2019roberta}}: is trained with dynamic masking, wherein the system learns to predict intentionally hidden sections of text within otherwise unannotated language examples. RoBERTa, implemented in PyTorch, modifies key hyperparameters in BERT, including removing BERT’s next-sentence pretraining objective and training with much larger mini-batches and learning rates.

\textbf{XLM-RoBERTa (XLM-R) \cite{conneau2019unsupervised}}: is a multilingual model trained using over two terabytes of cleaned and filtered CommonCrawl data. Upsampling low-resource languages during training and vocabulary generation, generating a more extensive shared vocabulary, and raising the overall model capacity to 550 million parameters are all important contributions of XLM-R.

\textbf{PhoBERT \cite{nguyen2020phobert}:} For Vietnamese, the SOTA method was first released and called PhoBERT by Nguyen et al. \cite{nguyen2020phobert} for solving Vietnamese NLP problems. PhoBERT is a pre-trained model, which has the same idea as RoBERTa, a replication study of BERT is released by Liu et al. \cite{liu2019roberta}, and there are modifications to suit Vietnamese.

\subsubsection{Combined approach}
BERT and CNN combined models have recently been widely used to classify short text collected from social networks, particularly to classify hate and offensive comments and achieves promising results \cite{safaya2020kuisail,liu2020hybrid,saha2021hate}. In this paper, variant BERT and CNN combined models are deployed to evaluate the efficiency of the combined models in classifying hate and offensive comments for Vietnamese. Furthermore, compared to BERT-CNN, RoBERTa-CNN, and XLMR-CNN models, our proposed PhoBERT-CNN model provides an insight into the effect of monolingual and multilingual pre-trained language models on this task.

\subsection{HSD Performance Evaluation Metric} \label{metric}
This section explain the evaluation metrics used in this paper. Accuracy and average macro F1-score are the popular and widely used metric for classification tasks in general and identifying hate and offensive comments in particular \cite{luu2021large,vu2020hsd,do2019hate,nguyen2019vais,van1991nlp,huu2019automated,luu2020comparison,huynh2020simple}. However, due to the significantly unbalanced classes in the given datasets, the average macro F1-score, which is the harmonic mean of Precision and Recall, is the most suitable metric for this task \cite{sigurbergsson2019offensive,chicco2020advantages}. As a result, when evaluating model performance, we opted to utilize the average macro F1-score (\%) as the primary metric and the Accuracy (\%) to provide additional information. Equation \ref{eqn:ac} to equation \ref{eqn:f1} presents the measures for multi-class classification for multi classes $C_i$, i $\in$ \text{\{1, 2, 3\}} (denoted by CLEAN, OFFENSIVE and HATE, respectively). Where $tp_i$ are true positive for $C_i$, and $fp_{i}$ – false positive, $fn_{i}$ – false negative, and $tn_{i}$ – true negative counts respectively. $M$ indices represent macro-averaging.

\begin{align}\label{eqn:ac}
Accuracy = \frac{\sum_{i=1}^{3} \frac{tp_i + tn_i}{tp_i +fp_i + tn_i + fn_i}}{3}
\end{align}

\begin{align}\label{eqn:pre}
Precision_{M} = \frac{\sum_{i=1}^{3} \frac{tp_i}{tp_i +fp_i}}{3}
\end{align}

\begin{align}\label{eqn:re}
Recall_{M} = \frac{\sum_{i=1}^{3} \frac{tp_i}{tp_i +fn_i}}{3}
\end{align}

\begin{align}\label{eqn:f1}
    F1 = 2 * \frac{\text{Precision$_M$ * Recall$_M$}}{\text{Precision$_M$ + Recall$_M$}}
\end{align}

\subsection{Vietnamese Hate Speech Detection Datasets}
\label{dataset}
We use the ViHSD dataset published by Luu et al. \cite{luu2021large} as the primary dataset to build the classification models. The ViHSD dataset \cite{luu2021large} consists of 33,400 comments collected from popular social networking sites in Vietnam such as Facebook and Youtube. This dataset is divided into training, development, test sets corresponding to a ratio of 7:1:2.

In addition, to illustrate the solution applicability and efficacy on the social media data domain, we also evaluate our solution on the HSD-VLSP dataset published by Vu et al. \cite{vu2020hsd}. HSD-VLSP is a Vietnamese Hate Speech Detection dataset on social-network comments provided by VSLP 2019 shared-task \cite{vu2020hsd}. This dataset contains 20,345 comments and posts on social networks.

Each comment in both datasets is assigned one of three labels: CLEAN, OFFENSIVE, or HATE. Table \ref{tab:ostatistics} shows overview statistics of the datasets. According to the statistics, there is a significant difference in the number of CLEAN comments compared to OFFENSIVE and HATE comments.

\begin{table*}[h]
\centering
\caption{Overview statistics of the two Vietnamese HSD datasets.}
\label{tab:ostatistics}
\resizebox{\textwidth}{!}{
\begin{tabular}{llrl}
\hline
\multicolumn{1}{c}{\textbf{Dataset}} & \multicolumn{1}{c}{\textbf{Labels}} & \multicolumn{1}{c}{\textbf{\begin{tabular}[c]{@{}c@{}}Percentage \\ (\%)\end{tabular}}} & \multicolumn{1}{c}{\textbf{Example}}                                                                              \\ \hline
\textbf{}                             & CLEAN                               & 82.71                                                                                   & \begin{tabular}[c]{@{}l@{}}link đâu thằng kia\\ (\textbf{English:} where is the link, man)\end{tabular}                         \\
\textbf{ViHSD}                        & OFFENSIVE                           & 6.77                                                                                    & \begin{tabular}[c]{@{}l@{}}vkl.\\ (\textbf{English:} cuss.)\end{tabular}                                                   \\
\textbf{}                             & HATE                                & 10.52                                                                                    & \begin{tabular}[c]{@{}l@{}}thầy đ*t mẹ giả tạo vl =))\\ (\textbf{English:} the teacher is so f*cking fake =))\end{tabular} \\ \hline
\textbf{}                             & CLEAN                               & 91.49                                                                                   & \begin{tabular}[c]{@{}l@{}}cho xíu nhạc đi\\ (\textbf{English:} some music please)\end{tabular}                           \\
\textbf{HSD-VLSP}                     & OFFENSIVE                           & 5.02                                                                                    & \begin{tabular}[c]{@{}l@{}}đ*o lấy vk nữa đâu\\ (\textbf{English:} no more f*cking married)\end{tabular}                 \\
                                      & HATE                                & 3.49                                 & \begin{tabular}[c]{@{}l@{}}thằng già ch* ch*t\\ (\textbf{English:} f*ck that old man)\end{tabular}                                       \\ \hline
\end{tabular}
}
\end{table*}

\begin{figure}[H]
\centering
    \begin{subfigure}[ht]{0.49\textwidth}
    \centering
    \includegraphics[width=\linewidth]{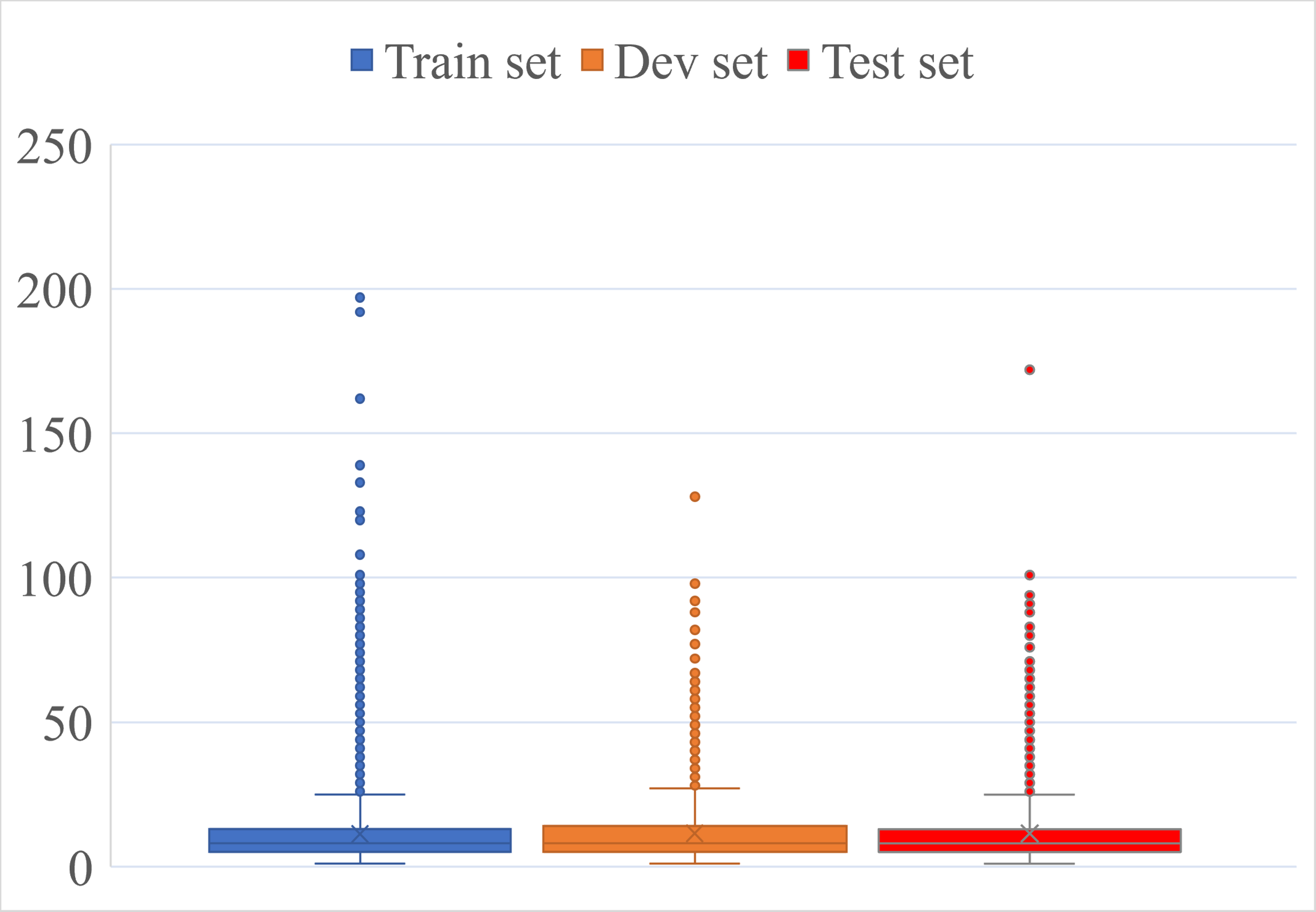}
    \caption{ViHSD dataset.}
    \label{fig:len_vihsd}
    \end{subfigure}
\hfill
    \begin{subfigure}[ht]{0.49\textwidth}
    \centering
    \includegraphics[width=\linewidth]{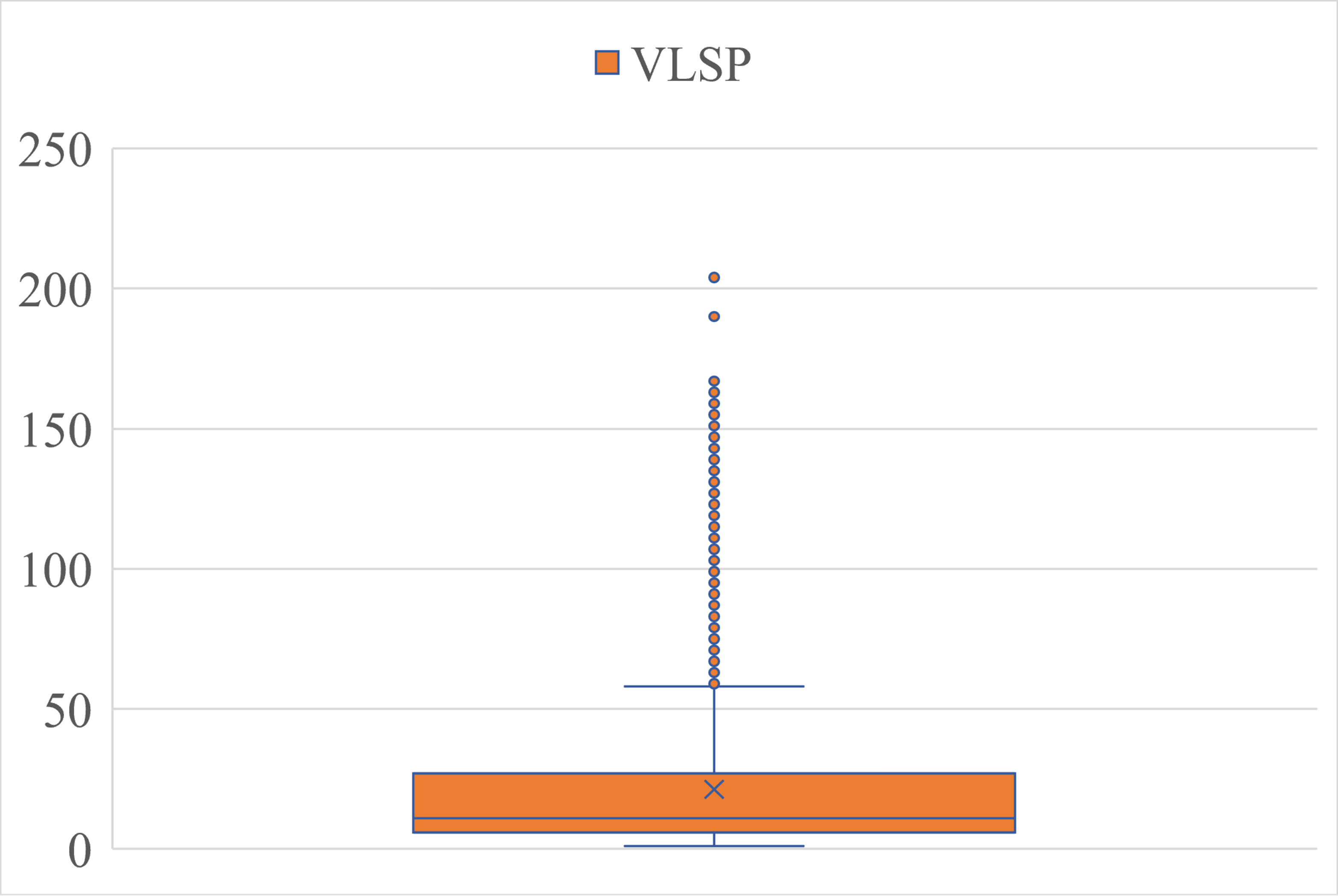}
    \caption{HSD-VLSP dataset.}
    \label{fig:len_hsd}
    \end{subfigure}
    \caption{Distribution of the comments length in the two Vietnamese HSD datasets.}
    \label{fig:data_lengths}
\end{figure}

The distribution of comment lengths in the ViHSD and HSD-VLSP datasets is depicted in Figure \ref{fig:data_lengths}. In the two datasets, ViHSD and HSD-VLSP, we can observe that the average length of the comments is 11.51 and 21.31 words, and the length of the comments is in the range (1; 25) and (1; 58.5), respectively. Besides, we found that comments are often short because users tend to be brief and use many acronyms. As demonstrated in Section \ref{phase2} and the statistics in Table \ref{tab:TeenStop_count}, one of the reasons for the relatively short average length of comments in the two datasets is the inclination of users to overuse acronyms, teen-codes to save time and type faster.

\subsection{Experimental Settings} \label{settings}
As we described in Section \ref{phuongphap}, we experimented on the given datasets with four approaches: traditional machine learning, deep learning, transfer learning, and the combined approach. Following the study of Son et al. \cite{luu2021large}, all findings from the ViHSD dataset are presented on the test set, while the development set is utilized for hyper-parameter tuning. Furthermore, due to the secrecy of HSD-VLSP, we employ KFoldCrossvalidation (k=5) for training and testing our models as a solution for the lack of a test set \cite{vu2020hsd,luu2020comparison,huynh2020simple}. 

Sections \ref{ml} and \ref{dl} provide the optimal hyper-parameters that we grid-searched for Machine Learning and Deep Learning models, respectively. On the other hand, we use an Adam optimizer with a fixed learning rate of 2e-5 and a batch size of 64 to fine-tune the hyperparameters of the PhoBERT-CNN model and the other baseline models.

\subsubsection{\textbf{Machine Learning approach:}}\label{ml} In this paper, we implement several machine learning models such as Multinomial Naive Bayes, Logistic Regression, Decision Tree, and Random Forest. In addition, we use the TF-IDF technique with parameter ngram\_range is (1,2) for feature extraction. Besides, we also use the class-weigh algorithm to solve the imbalance between three labels, but we do not get any better results.

•   \textbf{Multinomial Naive Bayes:} We use MutinomialNB with ``alpha'' = 1.0.

•   \textbf{Logistic Regression:} This model is implemented with the parameters
C = 1.0, solver = ``lbfgs'', maxIter = 20, and regParam = 0.3.

•   \textbf{Decision Tree:} Model parameters include n\_estimators = 108, random\_state = ``None'', class\_weight = ``balanced'', max\_depth = 17, and min\_samples\_leaf = 3.

•	\textbf{Random Forest:} We implement a Random Forest Classifier model with numTrees = 200, maxDepth = 10, maxBins = 64.

\subsubsection{\textbf{Deep Learning approach:}}\label{dl} This paper uses these two models: Text-CNN and Bi-LSTM. Otherwise, we also fine-tune those models with novel pre-trained word embeddings, which are Vietnamese word embedding ETNLP \cite{vu2019etnlp} and 300 dimensions with character n-grams PhoW2V: the word embedding used for Vietnamese and pre-trained Word2vec syllable \cite{phow2v_vitext2sql}.

•   \textbf{Text-CNN:} we set up four conv2D layers with 32 filters at sizes 1, 2, 3, 5 and used softmax for activation. In addition, we set batch size equal to 64, max sequence length is 40, and dropout is 0.4 for this model. 

•	\textbf{Bi-LSTM:} the structure of this model has a bidirectional layer followed by a max-pooling 1D, a dense layer has 50 in size for both  activation and softmax activation. We set batch size equal to 64, max sequence length is 68, and dropout is 0.4.

\subsubsection{\textbf{Transfer Learning approach:}} We implement BERT$_{large}$ cased \cite{devlin2018bert}, RoBERTa$_{large}$ \cite{liu2019roberta}, XLM-R$_{large}$ \cite{conneau2019unsupervised}, and PhoBERT$_{large}$ \cite{nguyen2020phobert} for this approach. They runs with their max sequence length is 60, batch size is 64, learning rate is at 2e-5, accumulation steps are 5, learning rate decay steps are 70.

\subsubsection{\textbf{Proposed approach (PhoBERT-CNN)}} In this approach, we combine the PhoBERT$_{large}$ pre-trained model \cite{nguyen2020phobert} from HuggingFace with the Text-CNN model. The output of PhoBERT$_{large}$ pre-trained is used as embedding input for the Text-CNN.

•	The PhoBERT$_{large}$ pre-trained is initialized with a max length is 20.

•	The Text-CNN is built with four layers of conv1D with filter size is 32 and size 1, 2, 3, 5, respectively.


\subsection{Analysis and Discussion of Experimental Results}
\label{results}
\subsubsection{Verifying the performance of the proposed PhoBERT-CNN model}
Table \ref{tab:results} shows our results from the experiments conducted. Experiment results show that the PhoBERT-CNN model outperforms traditional machine learning models on two benchmark datasets, ViHSD and HSD-VLSP, respectively, by an F1-score of 7.51$\pm$6.59\% and 16.25$\pm$13.89\%. With the deep learning approach, the Text-CNN model outperforms the Bi-LSTM model for the short text classification tasks in general and the HSD problem in particular. Among our single models, PhoBERT archives the highest results on the ViHSD \cite{luu2021large} and HSD-VLSP \cite{vu2020hsd} datasets. PhoBERT can perform parallel computations for words, reduce vanishing gradients, and help the model learn better. Our combined PhoBERT-CNN model outperforms the baseline models on the ViHSD dataset by 7.51$\pm$6.59\% in macro F1-score and 2.16$\pm$1.97\% in Accuracy, respectively. Besides, the proposed approach also demonstrates its efficacy in the Vietnamese social network data domain. On the HSD-VLSP dataset, PhoBERT-CNN achieves the best results, with a macro F1-score of 90.89\% and an Accuracy of 98.26\%. 
\begin{table}[ht]
\begin{center}
\begin{minipage}{\textwidth}
\caption{Evaluation results on the two Vietnamese HSD datasets.}
\label{tab:results}
\resizebox{\textwidth}{!}{
\begin{tabular}{lcccc}
\hline
\multirow{2}{*}{\textbf{Models}}          & \multicolumn{2}{c}{\textbf{ViHSD}}              & \multicolumn{2}{c}{\textbf{HSD-VLSP}}                                                   \\ \cline{2-5} 
                        & \textbf{F1-score} & \textbf{Accuracy} &\multicolumn{1}{c}{\textbf{F1-score}} &\multicolumn{1}{c}{\textbf{Accuracy}} \\ \hline
Multinomial Naive Bayes          & 50.33 & 85.23 & 63.06 & 92.45\\
Logistic Regression              & 56.77 & 86.61 & 66.41 & 94.29\\
Desion Tree                      & 55.68 & 83.38 & 60.75 & 91.84\\
Random Forest                    & 54.35 & 85.45 & 68.46 & 95.06\\ \hline
Text-CNN + $fastText$            & 61.67 & 86.98 & 85.76 & 97.14\\
Text-CNN + $PhoW2V_{syllable}$   & 62.49 & 86.89 & 86.52 & 97.22\\
Text-CNN + $PhoW2V_{word}$       & 63.01 & 86.11 & 85.36 & 97.11\\
Bi-LSTM + $fastText$             & 60.80 & 85.85 & 86.21 & 96.40\\
Bi-LSTM + $PhoW2V_{syllable}$    & 60.61 & 86.35 & 86.06 & 97.12\\
Bi-LSTM + $PhoW2V_{word}$        & 62.66 & 85.99 & 84.04 & 96.79 \\ \hline
BERT                             & 60.29 & 84.52 & 85.41 & 96.19\\
RoBERTa                          & 61.49 & 83.04 & 85.79 & 96.95\\
XLM-R                            & 62.38 & 83.62 & 86.57 & 97.15\\
PhoBERT                          & 63.51 & 87.13 & 86.68 & 97.58\\\hline
BERT - CNN                       & 61.26 & 85.90 & 86.37 & 96.17 \\
RoBERTa - CNN                    & 62.47 & 84.54 & 86.48 & 96.38 \\
XLMR - CNN                       & 63.34 & 85.48 & 88.53 & 96.92 \\\hline
\textbf{PhoBERT-CNN} & \textbf{64.43} & \textbf{87.17}                                          & \textbf{90.89} & \textbf{98.26}\\
\hline
\end{tabular}
}
\end{minipage}
\end{center}
\end{table}

On the other hand, the monolingual pre-trained language model for Vietnamese, particularly PhoBERT, outperforms the multilingual models on the task of Vietnamese HSD. Furthermore, combining the BERT and its variants such as PhoBERT, RoBERTa, and XLM-R with the CNN improves their performance by up to 0.98\% and 4.21\% macro F1-score in the two datasets ViHSD and HSD-VLSP, respectively. 

\subsubsection{Error analysis and Disscusion} \label{error}
\begin{figure}[H]
\centering
    \begin{subfigure}[ht]{0.49\textwidth}
    \centering
    \includegraphics[width=\linewidth]{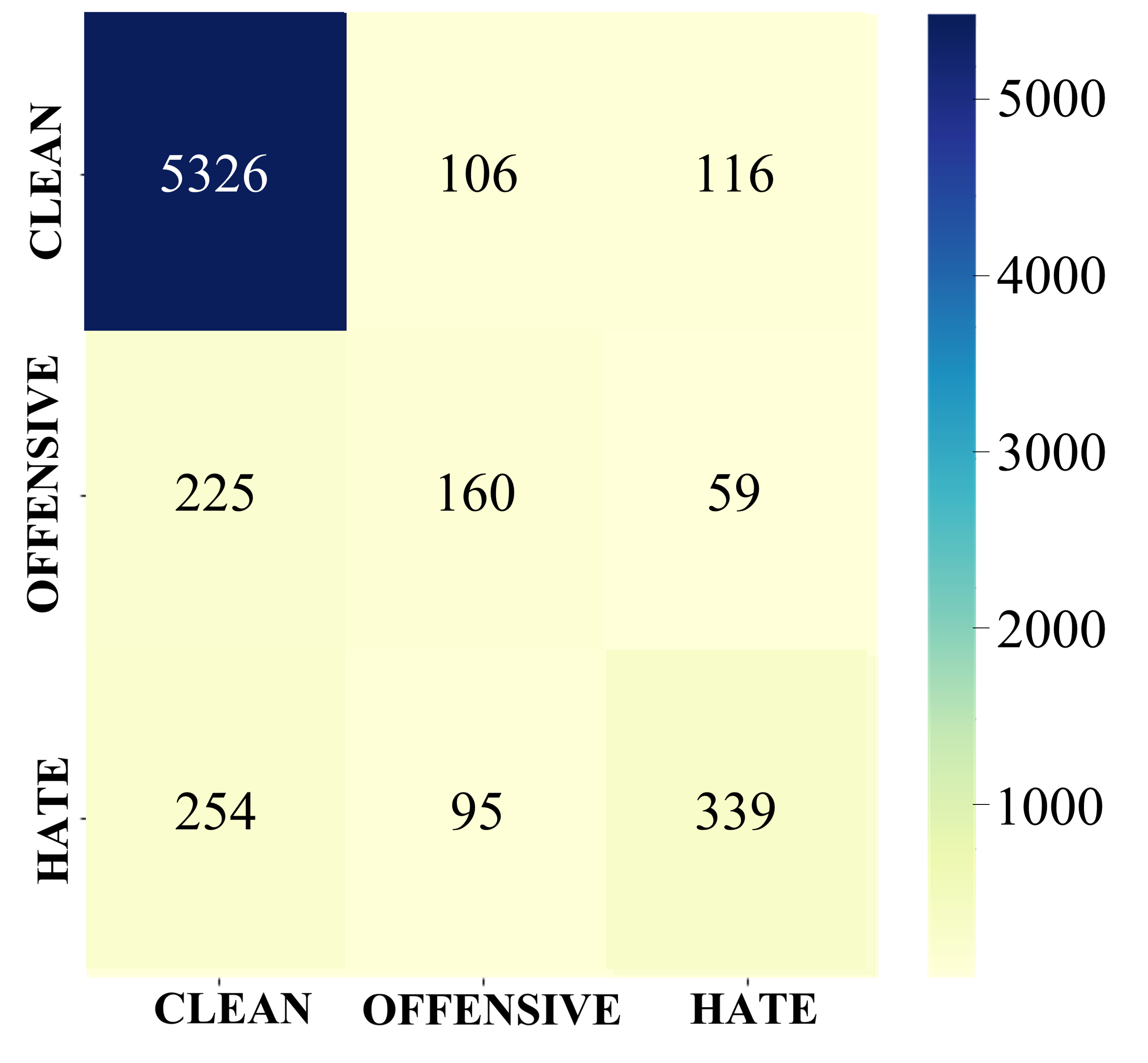}
    \caption{ViHSD dataset.}
    \label{fig:cfs1}
    \end{subfigure}
\hfill
    \begin{subfigure}[ht]{0.49\textwidth}
    \centering
    \includegraphics[width=\linewidth]{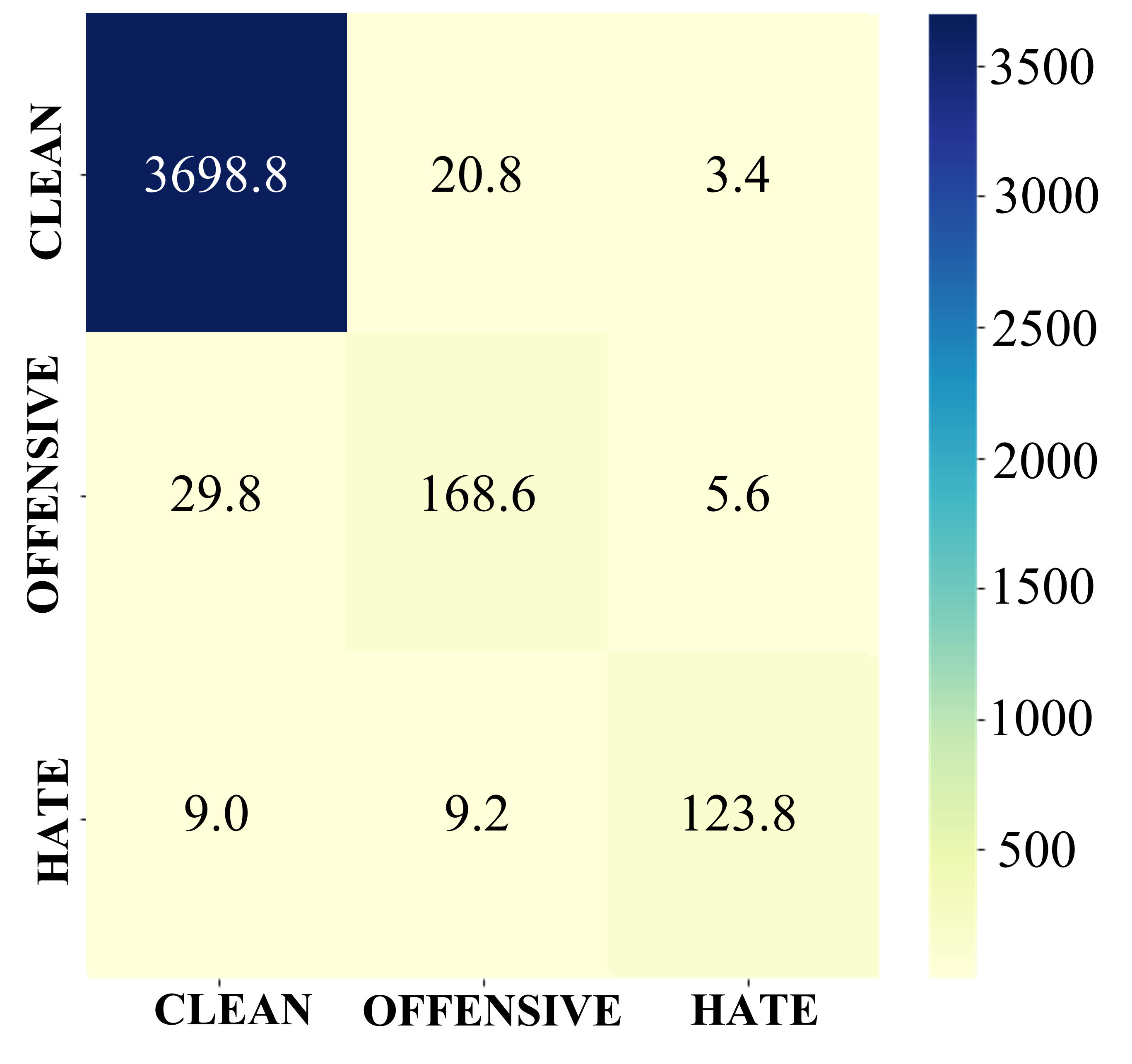}
    \caption{HSD-VLSP dataset.}
    \label{fig:cfs2}
    \end{subfigure}
    \caption{Confusion matrix of our proposed system for Vietnamese HSD.}
    \label{fig:cfs}
\end{figure}

The confusion matrix of our best-performance model, PhoBERT-CNN, is used for error analysis to analyze the errors encountered in our system. The confusion matrices of our best model when making predictions on the test set are shown in Figure \ref{fig:cfs}. As a result of the data imbalance, we observe that ability of our system to predict on the CLEAN label is better than the OFFENSIVE and HATE labels.

There are still some comments with misclassification in the dataset due to the ambiguity in identifying the labels. As described in Section \ref{phobertcnn}, the combination of the PhoBERT and the CNN contributes to improvement of the performance of the classifier by extracting the features related to the keywords representing the main idea of the sentence. However, CNN's extraction and learning of these keywords are sometimes too sensitive and subjective, leading to confusion in detecting the corresponding label of the comments. We can see that many misclassified comments are affected by the decision keywords, such as in the label CLEAN but are misclassified 
with the topic HATE due to the keyword ``lon\textsubscript{{\textit{big/pussy}}}'', ``nham\textsubscript{{\textit{miss/bullshit}}}'', ``đm\textsubscript{{\textit{fuck/fate}}}''. In addition, the HATE label and OFFENSIVE label is often confused with the CLEAN label because they contain racist or insinuating content that makes them challenging to predict \cite{luu2021large,davidson2017automated}.

\begin{table}[ht]
\begin{center}
\begin{minipage}{\textwidth}
\centering
\caption{Several examples of classification error on the given datasets.}
\begin{tabular}{p{6.8cm}p{1.8cm}p{1.8cm}}
\hline
\multicolumn{1}{l}{\textbf{Comment}}                                                                              & \makecell[c]{\textbf{Label}} & \makecell[c]{\textbf{Prediction}} \\ \hline
\begin{tabular}[c]{@{}l@{}}\makecell[l]{Đừng cố biện minh =)))) choi lon\\
(\textbf{English:} Do not try to make excuses, play big)}\end{tabular}          & \makecell[c]{CLEAN}               & \makecell[c]{HATE}               \\
\begin{tabular}[c]{@{}l@{}}\makecell[l]{Lắm tiền mà chả có nổi ý thức của loài người :))\\
(\textbf{English:} Money can not buy human consciousness :)))}\end{tabular}          & \makecell[c]{CLEAN}               & \makecell[c]{HATE}               \\
\begin{tabular}[c]{@{}l@{}}\makecell[l]{con này hết thuốc chữa rồi\\
(\textbf{English:} I am done with this dumb ass)}\end{tabular}                      & \makecell[c]{HATE}                & \makecell[c]{CLEAN}                 \\
\begin{tabular}[c]{@{}l@{}}\makecell[l]{Nham\\
(\textbf{English:} Bullshit)}\end{tabular}          & \makecell[c]{OFFENSIVE}               & \makecell[c]{HATE}               \\ \hline
\end{tabular}
\label{tab:erroranalysis}
\end{minipage}
\end{center}
\end{table}

\subsubsection{Augumentation data results}\label{aug_results}
One of the challenging problems in our task, as described in Section \ref{dataset}, is data imbalance, which has a negative impact on classification model performance. As a result, inspired by the work of Wei and Zou \cite{wei-zou-2019-eda}, we intend to apply EDA techniques to the ViHSD dataset and the HSD-VLSP dataset in this paper to deal with imbalanced data and improve the performance of classification models.

We used the data divided into training, development, and test sets by Luu et al. \cite{luu2021large} for the ViHSD dataset. The training set was used to implement EDA techniques, the development set to fine-tune classifier hyper-parameters, and the test set to evaluate our models.

For training and testing our models on the HSD-VLSP dataset, we employed five-fold cross-validation. Following the same approach as in the previous study \cite{luu2020comparison,luu-etal-2020-empirical}, we preserve the test set and augment the training set with EDA techniques for each fold.

Table \ref{tab:aug-compare} presents the results of a comparison of model performance with and without data augmentation techniques. The results indicate that using data augmentation approaches improves models performance by up to 9.34\% and 24.87\% macro F1-score in the two datasets, ViHSD and HSD-VLSP, respectively. Our proposed PhoBERT-CNN model, on the other hand, outperforms the baseline models 4.93$\pm$4.41\% and 16.76$\pm$8.11\% macro F1-score. According to the results, we conclude that using appropriate data augmentation techniques can significantly improve the performance of models. However, in some cases, data augmentation techniques could decrease model accuracy. As a result, it is critical to investigate if applying data augmentation techniques to a certain situation is appropriate.

\begin{table}[h]
\centering
\caption{The results compare the performance of two datasets, ViHSD and HSD-VLSP, for data augmentation techniques. ``W'' and ``W/o'' denote that the results are evaluated with data augmentation techniques and without data augmentation techniques.}
\label{tab:aug-compare}
\resizebox{\columnwidth}{!}{%
\begin{tabular}{lcccccccc}
\hline
\multirow{3}{*}{\textbf{Model}} &
  \multicolumn{4}{c}{\textbf{ViHSD}} &
  \multicolumn{4}{c}{\textbf{HSD-VLSP}} \\ \cline{2-9} 
 &
  \multicolumn{2}{c}{\textbf{F1-score}} &
  \multicolumn{2}{c}{\textbf{Accuracy}} &
  \multicolumn{2}{c}{\textbf{F1-score}} &
  \multicolumn{2}{c}{\textbf{Accuracy}} \\ \cline{2-9} 
 &
  \multicolumn{1}{c}{\textbf{W/o}} &
  \multicolumn{1}{c}{\textbf{W}} &
  \multicolumn{1}{c}{\textbf{W/o}} &
  \multicolumn{1}{c}{\textbf{W}} &
  \multicolumn{1}{c}{\textbf{W/o}} &
  \multicolumn{1}{c}{\textbf{W}} &
  \multicolumn{1}{c}{\textbf{W/o}} &
  \multicolumn{1}{c}{\textbf{W}} \\ \hline
Multinomial Naive Bayes     & 50.33 & 59.67 ($\uparrow$9.34) & 85.23 & 84.95 ($\downarrow$0.28) & 63.06 & 87.93 ($\uparrow$24.87) & 92.45 & 89.95 ($\downarrow$2.50)\\
Logistic Regression         & 56.77 & 61.37 ($\uparrow$4.60) & 86.61 & 88.33 ($\uparrow$1.72) & 66.41 & 91.91 ($\uparrow$25.50) & 94.29 & 93.65 ($\downarrow$0.64)\\
Decision Tree               & 55.68 & 59.66 ($\uparrow$3.98) & 83.38 & 83.77 ($\uparrow$0.39) & 60.75 & 89.85 ($\uparrow$29.10) & 91.84 & 91.54 ($\downarrow$0.30)\\
Random Forest               & 54.35 & 61.89 ($\uparrow$7.54) & 85.45 & 86.58 ($\uparrow$1.13) & 68.46 & 94.01 ($\uparrow$25.55) & 95.06 & 95.25 ($\uparrow$0.19)\\ \hline
TextCNN + $fastText$          & 61.67 & 62.54 ($\uparrow$0.87) & 86.98 & 86.07 ($\downarrow$0.91) & 85.76 & 98.16 ($\uparrow$12.40) & 97.14 & 98.29 ($\uparrow$1.15)\\
TextCNN + $PhoW2V_{syllable}$ & 62.49 & 63.57 ($\uparrow$1.08) & 86.89 & 85.14 ($\downarrow$1.75) & 86.52 & 98.03 ($\uparrow$11.51) & 97.22 & 98.16 ($\uparrow$0.94)\\
TextCNN + $PhoW2V_{word}$     & 63.01 & 63.53 ($\uparrow$0.52) & 86.11 & 86.06 ($\downarrow$0.05) & 85.36 & 98.19 ($\uparrow$12.83) & 97.11 & 98.23 ($\uparrow$1.12)\\
BiLSTM + $fastText$           & 60.80 & 62.85 ($\uparrow$2.05) & 85.85 & 86.99 ($\uparrow$1.14) & 86.21 & 97.56 ($\uparrow$11.35) & 96.40 & 97.92 ($\uparrow$1.52)\\
BiLSTM + $PhoW2V_{syllable}$  & 60.61 & 62.74 ($\uparrow$2.13) & 86.35 & 87.73 ($\uparrow$1.38) & 86.06 & 97.47 ($\uparrow$11.41) & 97.12 & 97.73 ($\uparrow$0.61)\\
BiLSTM + $PhoW2V_{word}$      & 62.66 & 63.68 ($\uparrow$1.02) & 85.99 & 86.57 ($\uparrow$0.58) & 84.04 & 97.56 ($\uparrow$13.52) & 96.79 & 97.74 ($\uparrow$0.95)\\ \hline
BERT                        & 60.29 & 63.04 ($\uparrow$2.75) & 84.52 & 81.39 ($\downarrow$3.13) & 85.41 & 96.67 ($\uparrow$11.26) & 96.19 & 98.01 ($\uparrow$1.82)\\
RoBERTa                     & 61.49 & 60.39 ($\uparrow$0.99) & 83.04 & 83.87 ($\uparrow$0.83)& 85.79 & 94.44 ($\uparrow$8.65) & 96.95 & 97.11 ($\uparrow$0.16)\\
XLM-R                        & 62.38 & 64.89 ($\uparrow$2.51) & 83.62 & 82.07 ($\downarrow$1.55)& 86.57 & 95.50 ($\uparrow$8.93) & 97.15 & 97.48 ($\uparrow$0.33)\\
PhoBERT                     & 63.51 & 65.07 ($\uparrow$1.56) & 87.13 & 89.59 ($\uparrow$2.46) & 86.68 & 97.01 ($\uparrow$10.33) & 97.58 & 98.55 ($\uparrow$0.97)\\ \hline
BERT-CNN               & 61.26 & 62.99 ($\uparrow$3.73) & 85.90 & 75.57 ($\downarrow$0.33) & 86.37 & 96.54 ($\uparrow$10.17) & 96.17 & 97.98 ($\uparrow$1.81)\\
RoBERTa-CNN                  & 62.47 & 64.02 ($\uparrow$1.55) & 84.54 & 88.06 ($\uparrow$3.52) & 86.48 & 95.47 ($\uparrow$8.99) & 96.38 & 98.01 ($\uparrow$1.63)\\
XLMR-CNN                  & 63.34 & 67.29 ($\uparrow$3.95) & 85.48 & 80.57 ($\downarrow$4.91) & 88.53 & 97.85 ($\uparrow$9.32)& 96.92 & 98.04 ($\uparrow$1.12)\\ \hline
\textbf{PhoBERT-CNN} &
  \textbf{64.43} &
  \textbf{67.46 ($\uparrow$3.03)} &
  \textbf{87.17} &
  \textbf{87.76 ($\uparrow$6.41)} &
  \textbf{90.89} &
  \textbf{98.45 ($\uparrow$7.56)} &
  \textbf{98.26} &
  \textbf{98.59 ($\uparrow$0.33)} \\ \hline
\end{tabular}%
}
\end{table}

\subsubsection{Comparison with Previous Studies}
\label{comparison}
We conducted experiments that compare the performance of our proposed data pre-processing techniques against the previous pre-processing, including comparing it against the Luu et al. \cite{luu2021large} and Huynh et al. \cite{huynh2020simple} studies. We choose to compare with the studies of Luu et al., and Huynh et al. \cite{huynh2020simple} because these are the works that implement effective pre-processing techniques and achieve more positive results than previous studies on the same dataset and evaluation measure \cite{luu2021large,huynh2020simple}. The experimental results are presented in Table \ref{tab:pre-compare}. The results show that our proposed pre-processing techniques significantly improve the performance of the PhoBERT-CNN model (increase by 4.80\% and 9.70\% macro F1-score on two datasets, ViHSD and HSD-VLSP). Moreover, our proposed pre-processing techniques improve the performance of the baseline models 2.82$\pm$2.23\% and 20.51$\pm$11.05\% macro F1-score on two datasets, ViHSD and HSD-VLSP, respectively. As a result, we conclude that the proposed pre-processing techniques are efficient in Vietnamese hate speech detection on social media. 

\begin{table}[h]
\centering
\caption{The comparison with previous pre-processing techniques on ViHSD test set and HSD-VLSP dataset. "LT," "HT," and "OT" refer to Luu et al. \cite{luu2021large}, Huynh et al. \cite{huynh2020simple}, and our proposed pre-processing techniques, respectively.} 
\label{tab:pre-compare}
\resizebox{\columnwidth}{!}{%
\begin{tabular}{lcccccccc}
\hline
\multirow{3}{*}{\textbf{Model}} &
  \multicolumn{4}{c}{\textbf{ViHSD}} &
  \multicolumn{4}{c}{\textbf{HSD-VLSP}} \\ \cline{2-9} 
 &
  \multicolumn{2}{c}{\textbf{F1-score}} &
  \multicolumn{2}{c}{\textbf{Accuracy}} &
  \multicolumn{2}{c}{\textbf{F1-score}} &
  \multicolumn{2}{c}{\textbf{Accuracy}} \\ \cline{2-9} 
 &
  \textbf{LT} &
  \textbf{OT} &
  \textbf{LT} &
  \textbf{OT} &
  \textbf{HT} &
  \textbf{OT} &
  \textbf{HT} &
  \textbf{OT} \\ \hline
Multinomial Naive Bayes     & 57.70 & 59.67 ($\uparrow$1.97) & 85.01 & 84.95 ($\downarrow$0.06) & 61.50 & 87.93 ($\uparrow$26.43) & 91.54 & 89.95 ($\downarrow$1.59)\\
Logistic Regression         & 54.53 & 61.37 ($\uparrow$6.84) & 86.27 & 88.33 ($\uparrow$2.06) & 63.98 & 91.91 ($\uparrow$27.93) & 94.01 & 93.65 ($\downarrow$0.36)\\
Decision Tree               & 55.48 & 59.66 ($\uparrow$4.18) & 83.48 & 83.77 ($\uparrow$0.29) & 59.08 & 89.85 ($\uparrow$30.77) & 91.10 & 91.54 ($\uparrow$1.44)\\
Random Forest               & 54.01 & 61.89 ($\uparrow$7.88)& 85.61 & 86.58 ($\uparrow$0.97) & 66.26 & 94.01 ($\uparrow$27.75) & 94.21 & 95.25 ($\uparrow$1.04)\\ \hline
TextCNN + Fasttext          & 61.95 & 62.54 ($\uparrow$0.59) & 86.89 & 86.07 ($\downarrow$0.82) & 85.02 & 98.16 ($\uparrow$13.14) & 97.06 & 98.29 ($\uparrow$1.23)\\
TextCNN + Pho2Vec\_syllable & 59.25 & 63.57 ($\uparrow$4.32) & 86.03 & 85.14 ($\downarrow$0.89) & 84.99 & 98.03 ($\uparrow$13.04) & 97.01 & 98.16 ($\uparrow$1.15)\\
TextCNN + Pho2Vec\_word     & 60.01 & 63.53 ($\uparrow$3.52) & 86.44 & 86.06 ($\uparrow$0.38) & 84.28 & 98.19 ($\uparrow$13.91) & 96.24 & 98.23 ($\uparrow$1.99)\\
BiLSTM + Fasttext           & 60.37 & 62.85 ($\uparrow$2.48) & 85.18 & 86.99 ($\uparrow$1.81) & 85.33 & 97.56 ($\uparrow$12.23) & 93.86 & 97.92 ($\uparrow$4.06)\\
BiLSTM + Pho2Vec\_syllable  & 60.27 & 62.74 ($\uparrow$2.47) & 84.42 & 87.73 ($\uparrow$3.31) & 82.12 & 97.47 ($\uparrow$15.35) & 95.02 & 97.73 ($\uparrow$2.71)\\
BiLSTM + Pho2Vec\_word      & 61.09 & 63.68 ($\uparrow$2.59) & 85.04 & 86.57 ($\uparrow$1.53) & 82.18 & 97.56 ($\uparrow$15.38) & 94.57 & 97.74 ($\uparrow$3.17)\\ \hline
BERT                        & 53.85 & 63.04 ($\uparrow$9.19) & 82.47 & 81.39 ($\downarrow$1.08) & 65.11 & 96.67 ($\uparrow$31.56) & 96.79 & 98.01 ($\uparrow$1.22)\\
RoBERTa                     & 55.34 & 60.39 ($\uparrow$5.05) & 83.54 & 83.87  ($\uparrow$0.33) & 84.98 & 94.44 ($\uparrow$9.46) & 96.01 & 97.11 ($\uparrow$1.10)\\
XLMR                        & 61.28 & 64.89 ($\uparrow$3.61) & 86.12 & 82.07 ($\downarrow$4.05) & 85.34 & 95.50 ($\uparrow$10.16) & 96.88 & 97.48 ($\uparrow$0.60)\\
PhoBERT                     & 60.58 & 65.07 ($\uparrow$4.49) & 86.84 & 89.59 ($\uparrow$2.75) & 86.02 & 97.01 ($\uparrow$10.99) & 96.58 & 98.55 ($\uparrow$1.97)\\ \hline
BERT - CNN                  & 59.81 & 64.02 ($\uparrow$4.21) & 84.57 & 88.06 ($\uparrow$3.49) & 85.01 & 95.47 ($\uparrow$10.46) & 95.17 & 98.01 ($\uparrow$2.84)\\
RoBERTa - CNN               & 58.28 & 62.99 ($\uparrow$4.71) & 84.01 & 75.57 ($\downarrow$8.44) & 85.11 & 96.54 ($\uparrow$11.43) & 95.55 & 97.98 ($\uparrow$2.43)\\
XLMR - CNN                  & 62.39 & 67.29 ($\uparrow$4.90) & 84.68 & 80.57 ($\downarrow$4.11) & 87.34 & 97.85 ($\uparrow$10.51) & 95.08 & 98.04 ($\uparrow$2.96)\\ \hline
\textbf{PhoBERT - CNN} &
  \textbf{62.66} &
  \textbf{67.46 ($\uparrow$4.80)} &
  \textbf{87.07} &
  \textbf{87.76 ($\uparrow$0.69)} &
  \textbf{88.75} &
  \textbf{98.45 ($\uparrow$9.70)} &
  \textbf{98.07} &
  \textbf{98.59 ($\uparrow$0.52)} \\ \hline
\end{tabular}%
}
\end{table}

Our approach outperforms previous studies on the ViHSD and HSD-VLSP datasets. The same evaluation metrics as previous studies are used to make fair comparisons. We utilize the average macro F1 (\%), Accuracy (\%) for ViHSD dataset \cite{luu2021large} and the average macro F1 score (\%) for HSD-VLSP dataset \cite{vu2020hsd}. The Table \ref{tab:comparision} and Table \ref{tab:comparision2} show the best results we achieved compared to previous studies. Our combined PhoBERT-CNN model outperforms baseline models of Luu et al. \cite{luu2021large} on the ViHSD dataset by 5.88$\pm$1.11\% macro F1-score and 1.615$\pm$0.735\% Accuracy, respectively. Furthermore, PhoBERT-CNN also achieves the best results, with a macro F1-score of 98.45\% on the HSD-VLSP dataset. 

\begin{table}[h]
\begin{center}
\begin{minipage}{\textwidth}
\centering
\caption{The comparison with previous studies on ViHSD dataset.}
\label{tab:comparision}
\begin{tabular}{lrrrr}
\hline
\multicolumn{1}{c}{\textbf{Model}} & \textbf{F1-score} & \textbf{Accuracy} \\ \hline
GRU + fastText \cite{luu2021large}                   & 60.47             & 85.41             \\
Text-CNN + fastText \cite{luu2021large}                & 61.11             & 86.69             \\
XLM-R \cite{luu2021large}                            & 61.28             & 86.12             \\
DistilBERT \cite{luu2021large}                        & 62.42             & 86.22             \\ 
BERT \cite{luu2021large}                              &62.69    & 86.88    \\\hline
\textbf{Our approach (PhoBERT-CNN)}              & \textbf{67.46}         & \textbf{87.76}         \\ \hline
\end{tabular}
\end{minipage}
\end{center}
\end{table}

\begin{table}[h]
\centering
\caption{The comparison with previous studies on HSD-VLSP dataset. ${\text{}}^{*}$Indicates that the result is evaluated on a test set of the VLSP shared task 2019. Others use K-fold cross-validation to evaluate the model (k=5) following the study.}
\label{tab:comparision2}
\begin{tabular}{lr}
\hline
\textbf{Model} & \multicolumn{1}{l}{\textbf{F1-score}}\\ \hline
${\text{Bi-LSTM}}^{*}$ \cite{do2019hate} & 56.28\\
${\text{DCNN, Text-CNN, LSTM, LSTMCNN, SARNN}}^{*}$ \cite{nguyen2019vais}                                     & 58.45 \\
${\text{Logistic regression + Random Forest + Extra Tree}}^{*}$ \cite{van1991nlp}                              & 58.88 \\
${\text{Logistic regression}}^{*}$ \cite{huu2019automated}                                   & 61.97\\
Text-CNN \cite{luu2020comparison}      & 83.04\\
CNN + Bi-LSTM + LSTM \cite{huynh2020simple}                                                                       & \textbf{86.96}                                                                              \\ \hline
\textbf{Our approach (PhoBERT-CNN)}                       & \textbf{98.45}  \\ \hline
\end{tabular}
\end{table}

\subsubsection{Ablation Analysis of Proposed Method}
\label{ablationanalysis}
Our proposed approach is significantly more straightforward and efficient than most existing Vietnamese hate speech detection approaches (see Subsection \ref{comparison}). An ablation analysis was performed on the proposed approach to demonstrate the efficacy and alignment of the modules. Table \ref{tab:abaltion} shows that each module contributes to the overall performance of our approach. Without the pre-trained PhoBERT model, our method only achieved 63.54\% and 98.19\% macro F1-score on the ViHSD and HSD-VLSP datasets, respectively. Likewise, the inclusion of a combined model contributes to a significant improvement in approach performance up to 11.77\% in macro F1-score.

We have also compared our proposed pre-processing techniques with the pre-processing techniques of previous works \cite{huynh2020simple}. The Easy Data Augmentation (EDA) techniques \cite{wei-zou-2019-eda} was chosen since it is one of the novel solutions and has been shown to be successful when applied to sentiment analysis tasks, including the Vietnamese HSD task \cite{luu-etal-2020-empirical}. The results show that the EDA techniques significantly improve the performance of the PhoBERT-CNN model (increase by 6.01\% and 13.84\% macro F1-score in the two datasets, ViHSD and HSD-VLSP). These results verify the need for data augmentation for the Vietnamese HSD problem.

Moreover, removing pre-processing and pre-trained PhoBERT or Text-CNN models reduces performance dramatically. The results demonstrate the importance of data pre-processing in general, as well as the effectiveness of the pre-processing techniques we applied to identify Vietnamese HSD in particular. Combined models, especially pre-trained models combined with deep learning networks, could yield promising outcomes for improving performance in further study. As a result, we conclude that all proposed modules are crucial in Vietnamese hate speech detection on social media. 

\begin{table}[h]
\centering
\caption{Ablation test on our proposed approach. ``OP'', ``DA'', ``PB'', and ``TC'' denote the use of our proposed pre-processing, the data augumentation techniques, the pre-trained PhoBERT$_{large}$ model, and the Text-CNN model, respectively.}
\label{tab:abaltion}
\resizebox{\columnwidth}{!}{%
\begin{tabular}{cccccccc}
\hline
\multirow{2}{*}{\textbf{OP}} &
  \multirow{2}{*}{\textbf{DA}} &
  \multirow{2}{*}{\textbf{PB}} &
  \multirow{2}{*}{\textbf{TC}} &
  \multicolumn{2}{c}{\textbf{ViHSD}} &
  \multicolumn{2}{c}{\textbf{HSD-VLSP}} \\ \cline{5-8} 
  &   &   &   & \textbf{F1-score} & \textbf{Accuracy} & \textbf{F1-score} & \textbf{Accuracy} \\ \hline
\cmark & \cmark & \cmark & \cmark & \textbf{67.46}    & \textbf{87.76}    & \textbf{98.45}    & \textbf{98.59}    \\ \hdashline
\cmark & \cmark & \cmark & \xmark & 65.07 ($\downarrow$2.39) & 87.59 ($\downarrow$0.17) & 97.01 ($\downarrow$1.44) & 98.55 ($\downarrow$0.04)\\ \hdashline
\cmark & \cmark & \xmark & \cmark & 63.54 ($\downarrow$3.92) & 86.06 ($\downarrow$1.70) & 98.19 ($\downarrow$0.26) & 98.23 ($\downarrow$0.36)\\ \hdashline
\cmark & \xmark & \cmark & \cmark & 64.43 ($\downarrow$3.03) & 87.17 ($\downarrow$0.59) & 90.89 ($\downarrow$7.56) & 98.26 ($\downarrow$0.33)\\ \hdashline
\cmark & \xmark & \cmark & \xmark & 63.51 ($\downarrow$3.95) & 87.13 ($\downarrow$0.63) & 86.68 ($\downarrow$11.77) & 97.58 ($\downarrow$1.01)\\ \hdashline
\cmark & \xmark & \xmark & \cmark & 63.01 ($\downarrow$4.45) & 86.11 ($\downarrow$1.65)& 85.36 ($\downarrow$13.09) & 97.11 ($\downarrow$1.48)\\ \hdashline
\xmark & \cmark & \cmark & \cmark & 62.91 ($\downarrow$4.55) & 86.41 ($\downarrow$1.35) & 88.75 ($\downarrow$9.70) & 98.07 ($\downarrow$0.52)\\ \hdashline
\xmark & \cmark & \cmark & \xmark & 62.08 ($\downarrow$5.38) & 86.38 ($\downarrow$1.38) & 86.02 ($\downarrow$12.43) & 96.88 ($\downarrow$1.71)\\ \hdashline
\xmark & \cmark & \xmark & \cmark & 61.86 ($\downarrow$5.60) & 85.71 ($\downarrow$2.05) & 85.99 ($\downarrow$12.46) & 96.24 ($\downarrow$2.35)\\ \hdashline
\xmark & \xmark & \cmark & \cmark & 61.45 ($\downarrow$6.01) & 85.25 ($\downarrow$2.51) & 84.61 ($\downarrow$13.84) & 97.85 ($\downarrow$0.74)\\ \hdashline
\xmark & \xmark & \cmark & \xmark & 58.87 ($\downarrow$8.59) & 86.34 ($\downarrow$1.42) & 84.52 ($\downarrow$13.93) & 96.65 ($\downarrow$1.94)\\ \hdashline
\xmark & \xmark & \xmark & \cmark & 56.03 ($\downarrow$11.43)& 86.54 ($\downarrow$1.22) & 83.46 ($\downarrow$14.99) & 96.33 ($\downarrow$2.26)\\ \hline
\end{tabular}%
}
\end{table}

\subsubsection{Hate speech detection application with streaming data}
\label{sparkstreaming}

This section runs a pilot experiment on a system with a NVIDIA GEFORCE GTX 1650Ti GPU, an Intel i7-9750H CPU, and 16 GB of RAM. This experiment is conducted to analyze the effectiveness of our proposed system when deployed to handle streaming for a social networking platform in practice. We chose Youtube to conduct the pilot experiment because it is one of the largest social networking sites in Vietnam currently \cite{digital2020vietnam}. Besides, Youtube is the primary data source for constructing both the ViHSD \cite{luu2021large} and HSD-VLSP \cite{vu2020hsd} datasets. The successful construction and application of the system to the Youtube platform demonstrates the value of our study. Furthermore, our proposed system lays the groundwork for future research into HSD application systems for social networks.

At the online stage, the results are evaluated by verifying the prediction results returned by the system. Firstly, we randomly collect 500 comments with real-time access from Youtube via Youtube Data API. The collected comments will be stored in JSON format. Collected comments have many attributes and mainly include user-based and comment-based properties. However, we only focus on the comment-based description to classify the polarity labels that the comment conveys.

\begin{figure}[H]
    \centering
    \includegraphics[width=\linewidth]{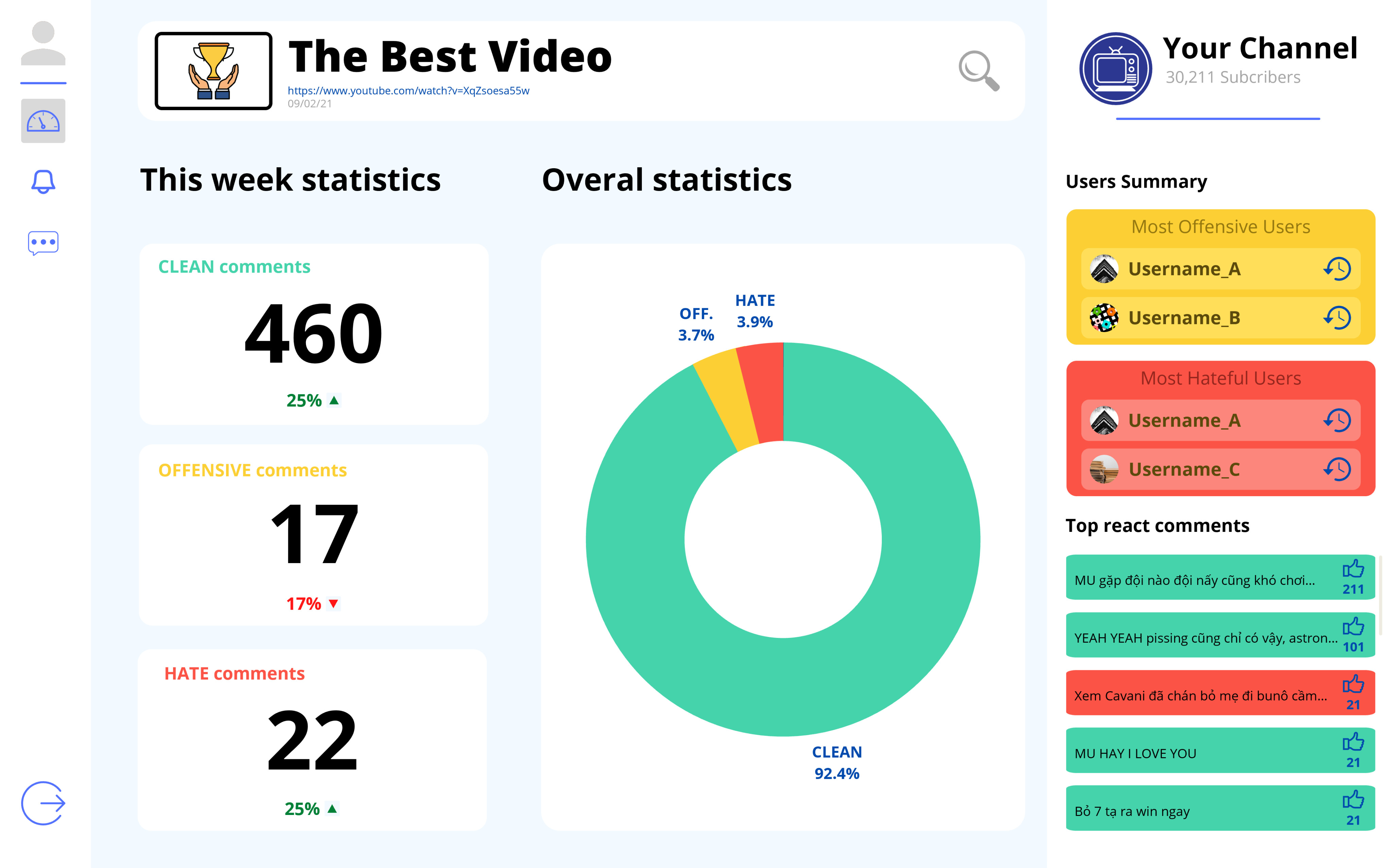}
    \caption{The interface of the HSD application with streaming data.}
    \label{fig:stream_result}
\end{figure}

Next, we hired three annotators with extensive experience constructing NLP datasets for Vietnamese, particularly ViHSD datasets, to annotate the collected comments. Three annotators independently label 500 comments with an inter-annotator agreement score of Cohen Kappa \cite{cohen1960coefficient} at K = 0.64. Figure \ref{fig:stream_result} shows an overview of statistics and the user interface of our application.

The results of our Hate speech detection application with streaming data are depicted as follows. The F1-score macro and Accuracy metrics are used to evaluate the accuracy of the system. With 500 test data points, our proposed system achieves an F1-score and Accuracy of 58.19\% and 82.02\%, respectively. We use Apache Spark Structured Streaming 3.1.1\footnote{\url{https://spark.apache.org/docs/3.1.1}} \cite{zaharia2016apache} in this article. This version includes Continuity Processing for Structured Streaming, which can provide low-latency responses on the order of 1ms, as opposed to the 1000ms that could receive with micro-batching in previous versions or similar frameworks. According to measurement findings, the proposed system processes and classifies a comment in an average of 1.56 seconds. However, due to limited hardware resources, our system's response time is still modest. Briefly, these research findings could serve as the foundation for further development and comparison with proposed solutions in the field.

\section{Conclusion and Future Works}
\label{ketluanvahuongphattrien}
This research proposes a novel state-of-the-art solution to the Vietnamese HSD task. We introduced a new data processing process with two phases to clean the given dataset well. The results show that our proposed pre-processing techniques significantly improve the performance of the PhoBERT-CNN model (increase by 4.80\% and 9.70\% macro F1-score on two datasets, ViHSD and HSD-VLSP). Moreover, our proposed pre-processing techniques improve the performance of the baseline models 3.58$\pm$3.38\% and 10.16$\pm$10.14\% macro F1-score on two datasets, ViHSD and HSD-VLSP, respectively. We were also successful in addressing the problems of data imbalance by applying the EDA techniques. Meanwhile, an efficient and straightforward approach for Vietnamese hate speech detection based on the combined model PhoBERT-CNN was proposed. The proposed combined model outperformed the baseline models and previous studies on the same dataset and evaluation measure. We achieved F1-score results on the ViHSD and HSD-VLSP datasets of 67.46\% and 98.45\%, respectively. Furthermore, we have successfully built a real-time hate speech detection system for Vietnamese using Spark streaming. The results obtained are pretty optimistic and reliable for solving the task, helping to reduce the occurrence of hate or offensive comments, and building a healthy and safe environment. 

Although the proposed system is still in its early phases, its application reach is vast. The proposed system can be applied to online newspapers or websites with low comment volume but need high censoring, as well as large social networks or forums. It creates a positive, civilized environment or satisfies the need to orient and safeguard vulnerable subjects such as the elderly and children. The application also serves as a foundation for agencies and organizations to review and monitor for management, psychological research, teaching, and other purposes.

Inspired by the success of Mozafari et al. \cite{mozafari2019bert}, we intend to implement more Vietnamese pre-trained language models to find a better model that achieves better performance in the Hate Speech Detection task. Moreover, our research lays the groundwork for future research in areas such as: (1) detecting multiple aspects and human rationales of hate speech \cite{mathew2020hatexplain}; (2) detecting hate and offensive spans at the word and phrase levels \cite{pavlopoulos2021semeval}.

\section*{Declarations}

\textbf{Conflict of interest} The authors declare that they have no conflict of interest.


\bibliography{sn-bibliography}

\end{document}